\documentclass[sn-mathphys]{sn-jnl}
\usepackage[utf8]{inputenc}
\usepackage{colortbl} 
\usepackage{hyperref}
\usepackage{amssymb} 
\usepackage{graphicx}
\usepackage{enumitem}
\usepackage{comment}
\usepackage{algpseudocode}

\usepackage{smartdiagram}
\usepackage{hhline}
\usepackage{float} 
\usepackage{makecell} 
\usepackage{amsmath}
\usepackage{cleveref} 

\usepackage[normalem]{ulem}

\begin{document}

\def\N{\mathbb{N}}
\def\R{\mathbb{R}}

\def\nn{\nonumber}

\newcommand{\bfast}{BFAST}
\newcommand{\arima}{ARIMA}
\newcommand{\cusum}{CUSUM}
\newcommand{\lstm}{LSTM}
\newcommand{\pandc}{P\&C}
\newcommand{\fpc}{Fpc}
\newcommand{\arlp}{ArlP}
\newcommand{\ocd}{OCD}

\title{Predictive change point detection for heterogeneous data}
\author*[1]{\fnm{Anna-Christina} \sur{Glock}}\email{\fontsize{10}{11}\selectfont anna-christina.glock@scch.at}

\author[1]{\fnm{Florian} \sur{Sobieczky}}\email{\fontsize{10}{11}\selectfont florian.sobieczky@scch.at}

\author[2]{\fnm{Johannes} \sur{Fürnkranz}}\email{\fontsize{10}{11}\selectfont juffi@faw.jku.at}

\author[3]{\fnm{Peter} \sur{Filzmoser}}\email{\fontsize{10}{11}\selectfont peter.filzmoser@tuwien.ac.at}

\author[4]{\fnm{Martin} \sur{Jech}}\email{\fontsize{10}{11}\selectfont martin.jech@ac2t.at}
\date{April 2023}

\affil*[1]{\fontsize{8}{9}\selectfont\orgname{Software Competence Center Hagenberg GmbH}, \orgaddress{\street{Softwarepark 32a}, \city{Hagenberg}, \postcode{4232}, \country{Austria}}}

\affil[2]{\fontsize{8}{9}\selectfont\orgdiv{Institute for Application Oriented Knowledge Processing (FAW)}, \orgname{Johannes Kepler University Linz}, \orgaddress{\street{Altenberger Straße 66B}, \city{Linz}, \postcode{4040}, \country{Austria}}}

\affil[3]{\fontsize{8}{9}\selectfont\orgdiv{Computational Statistics
Institute of Statistics and Mathematical Methods in Economics}, \orgname{Vienna University of Technology}, \orgaddress{\street{Wiedner Hauptstrasse 8-10}, \city{Vienna}, \postcode{1040}, \country{Austria}}}

\affil[4]{\fontsize{8}{9}\selectfont\orgname{AC2T research GmbH}, \orgaddress{\street{Hafenstraße 47-51}, \city{Linz}, \postcode{4020}, \country{Austria}}}

\abstract{ 
    An unsupervised change point detection (CPD) framework assisted by a predictive machine learning model called ''Predict and Compare'' is introduced which is able to detect change points online  under the presence of non-trivial trend patterns which must be prevented from triggering false positives. Different  predictive models for the required time series forecasting (Predict) step together with different statistical tests for deciding about the proximity of predicted and actual data (Compare step)  are allowed. Its performance is shown  for the Predict step being carried out by either an LSTM recursive neural network or an ARIMA linear time series model together with the CUSUM rule as Compare step method. It shows to perform best  in  comparison  to several other online CPD  methods for  detect  times  in the regime of low  numbers  of false positive detections. The  method's good performance is based on its ability to detect  structural changes in the presence complex  underlying trend patterns. The use case concerns tribological wear for which change points separating the run-in, steady-state, and divergent wear phases are detected. 
}


\maketitle

\section{Introduction}
\label{sec:introduction}

\subsection{The problem of change point detection}

Change point detection (CPD) in time series classically refers to analyzing the observed data in order to identify abrupt changes in the underlying latent probability distribution \cite{Lai_95,Wu_05,odile_2018}. A classical approach in this area is \mbox{\cusum \cite{pageContinuousInspectionSchemes1954a}}, which sequentially tracks a cumulative sum and flags a distribution change when the value exceeds a threshold determined by a 
sequential test \cite{wald}. This has been shown to be optimal in the sense of smallest detection times under the given expected length between false positives for asymptotically large average in-control run-lengths \cite{lord_1977}, and exact (finite sample) optimality in a decision-theoretic (mini-max) sense \cite{mous_2004,ritov,aue}. The number of fields in which this detector finds applications is large (e.g. medical \cite{Yang_Heart}, micro-economical \cite{harrison_Demand}, portfolio managerial \cite{manner_Copula}; see \cite{burgEvaluationChangePoint2020} for a review).

A major distinction between change point detection techniques is whether change points (CPs) are determined after a batch sample has been obtained ({\em offline} mode) or whether they are continuously updated each time a new sample point is added ({\em online} mode). The latter is setting the scene for sequential tests and is therefore often referred to as {\em sequential} change point detection \cite[p.~4]{amin_2017}\cite{siegmund}). The majority of CPD methods are offline \cite{truong}, as they have a wider range of applicability due to the additional range of available data after each proposed change point (unknown post-change parameters \cite[Chap.~1.1]{cao}). The availability of the entire data set is also useful for increasing the power of the tests. 
On the other hand, several types of processes involving the monitoring of sensor values are inherently online and therefore, do not allow the use of offline techniques (e.g. quality management in production \cite{hawkinsCumulativeSumCharts1998,lim}). In  \cite{amin_2017}, Sect. 4.2, it is recognized that each online CPD algorithm employs a sliding time-interval of a certain size within which the decision about the presence of a CP is made. Furthermore, the use of additional 'retrospective' windows characteristically belongs to several online CPD methods \cite{amin_2017}. Namely, they are those belonging to Bayesian modeling (e.g. \cite{pan_2005}), and Gaussian Process modeling \cite{huwel_dynamically_2022}. Several online CPD algorithms, however, just compare the data on the two sliding windows, instead of using a prediction. Comparison between different distributions, such as by the Mann-Whittney test, has recently been employed for developing a particle filter method for Capacity Reaction Point detection \cite{ma_rul_2021}. While this method belongs to the offline circle of CP detectors, there is a continuous transition from completely offline to completely online algorithms, and applications of particle filters may also find use in online techniques for small (tolerable) sliding window sizes.

It is precisely the sequential tests \cite{wald,siegmund} such as \cusum, and Exponentially Weighted Moving Average (EWMA) models which are exploited in applications (e.g., production process quality control \cite[Chap.~9]{SQC}) and which allow a rigorous {\em online} change point monitoring within a test-theoretic setting \cite[Chapter~1.4]{cao}. As noted in \cite[p.~22]{amin_2017}, defining online CPD methods for non-stationary data is an ongoing challenge.

One important generalization away from the step-like change points consists of gradually changing location parameters \cite{buecher,vogt,aueSteinbach}, called {\em gradual change points}. Already in \cite[Chap~1.5]{woodallSTATISTICALDESIGNCUSUM1993} it is pointed out, however, that test results for linear changes as opposed to step anomalies are not easy to distinguish in practice. Nevertheless, modeling gradual changes by linear transitions can already improve the characteristic average run length estimates of the \cusum{} process \cite{bissell}. A much more general assumption of the departure from step-anomalies is to suppose that stationarity only exists locally, for which \cite{vogt} have shown the advantages of a 'refined \cusum{} rule'. 

A further generalization consists of dropping the assumption of zero mean stationarity between the change points. In \cite[Sec.~2.2]{aue}, the possibility to assume, more generally, the validity of a linear regression model between two change points is exploited. Two interesting methods to include such linear trends between change points are the Continuous-piecewise-linear Pruned Optimal Partitioning (CPOP) \cite{fearnhead_cpop_2019} and Break detection for Additive Season and Trend (\mbox{\bfast{}) \cite{verbesselt_detecting_2010}}. The latter works also in the online case. 
Another successful method for online CPD for non-stationary data is by using a Bayesian approach \cite{AdamsMacKay}, as discussed in \cite[Chapter~C.1]{burgEvaluationChangePoint2020}, which also allows non-linear trends between change points. The online changepoint detection (\ocd) by Chen et al. \cite{chen_OCD_2022} is a very recent online detection method that uses aggregated likelihood test statistics for detection. Together with the classical \cusum{} rule we choose these methods as our benchmark references because they cover a wide methodological scope. We note that further recent approaches challenging the intermediate stationarity condition include optimization \cite{cao}, and other Bayesian methods \cite{agudelo,wang}, also admitting non-trivial 
intermediate trends.

\subsection{The predict-and-compare approach}

In this paper, we introduce \emph{Predict and Compare}, a novel CPD framework that is able to detect change points in non-trivial trend patterns, using predictive machine learning for modeling the trend, and the discrepancy between the forecasts and the observed data for detecting change points. This approach allows us to address the problem of CPD with an emphasis on

\begin{enumerate}[label=\emph{\Alph*}.]
    \item defining an {\bf online} change point detector,
    \item searching for CPs belonging to {\bf gradual} changes with few false positives,
    \item allowing for the presence of changing, non-trivial {\bf trends} between CPs.\\
\end{enumerate}

\begin{figure}[htb]{\tiny Transition between noisy (exponential) Run-in and (linear) Steady State increasing signal}
    \centering
    \includegraphics[width=1.0\linewidth]{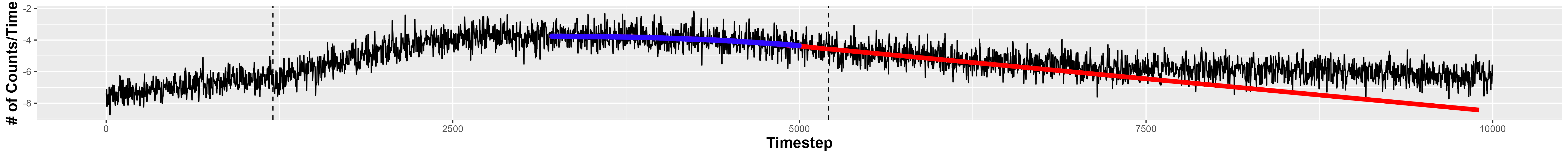}
    \caption{Principle of \pandc: The diagram shows CPs at the vertical dotted lines - it is standardized data from a tribological experiment about the wear occurring in a bearing (Sect. \ref{subsec:03_zScore}). From input data (blue) up to $t=5000$, an online prediction (red) is made starting at $t=5001$ and deviating from the trend of the data after the CP. This facilitates its subsequent detection by an online sequential statistical test. Note the data is heterogeneous, as different non-stationary trend patterns are concatenated.}
    \label{fig:principle}
    \end{figure}

Online detectors often focus on deviations from a stationary sequence of observed in-control input data. Our approach, however, allows detecting the CP in the presence of a non-trivial trend, possibly changing at the CP. We call this \emph{heterogeneous} data. The condition that the trend is not confused with the CP is made possible by it having previously been learned in the training process of the predictive model. The key  idea and contribution  of Predict and Compare (\pandc) is the ability to differentiate between true CPs and patterns belonging to previously learned regular (‘in-control’) trend patterns. These patterns may be much more complicated than linear patterns conventionally detected by CPD methods such as ARIMA or BFAST. An example consists of the important run-in processes characterized by their gradually dying out curvature. (\Cref{fig:principle} shows an example of such a transition from the tribological data of wear of a cylindrical bearing analyzed in \Cref{sec:exp_disc}).

While linear, damped, or seasonal trends are among the classically detectable patterns in time series analysis (see e.g. \cite{verbesselt_detecting_2010}, \cite[Sect. 2.3]{gardner85}), trends following a less obvious pattern which can only be detected by non-linear predictive models are mistaken as CPs by conventional CPD methods and typically leads to a high false positive detection rate specifically for gradual CPs. \pandc\, detects CPs only if they do not match predictions of the possibly complicated trends learned by a predictive model. This feature is responsible for a significant reduction in its false positive detection rate (Table 4).

\vspace{-0.5cm}    
The technical implementation of this idea first involves a 'historical' data set including the trend (abbrev. by $f$) from the 'distant past' for training a model $\widehat{f}$ estimating $f$. The model $\widehat{f}$ takes a finite sequence as input data from a {\em input window} representing the immediate past before the current point in time and predicts the sequence of values on a time window immediately behind the current moment (called the {\em prediction window}).  Thus, $\widehat{f}$ aims at capturing the general trend in the underlying probability distribution, so that significant (onsets of) changes from these predictions can be considered to be one of the CPs. \Cref{fig:principle}{} shows this principle idea for the CP between an exponential run-in and a noisy linearly increasing process (\Cref{tab:04_bestResForK}{} gives the results for comparable CPs on real tribological data).

The learning model involved in \pandc\, is trained to predict how the trend of the sequence of the time series progresses (Predict Step). For example, the run-in pattern typically observed after re-initiating a monitored production process is neither linear, nor periodic (see \Cref{fig:principle}). However, it is well predictable given only the onset of the run-in pattern from the input window. This is possible as long as it has been trained on a data set that includes sufficiently many run-in sub-processes of length fitting into the training window. The length of the training window thus restricts the recognizable pattern types to a certain maximal time scale. In \pandc\, the prediction estimate on the prediction window is then compared to the actual data (Compare Step) to check for CPs present {\em in addition} to the predicted run-in pattern.

By heterogeneous data, we mean that the process between CPs (i) may be non-stationary, and (ii) may change its pattern at the CPs (cf. \cite{krause})). As long as $\widehat{f}$ has been trained to recognize the occurring trend patterns as characteristics of the time series, \pandc\; allows recognizing them as 'normal data' from which changes are recognized as CPs: What is normal in the case of \pandc\, is only defined by what the predictive model has been trained to recognize as such. As there is no need for labels, in this way, \pandc\, belongs to the unsupervised method types (see \cite{amin_2017}, Sect. \ref{subsec:03_zScore}).

\subsection{Summary of contributions and overview}

The main contributions of this work can be summarized as follows:
\begin{itemize}
    \item We formally define Predict and Compare (\pandc{}), a general framework for on-line change point detection that can be instantiated with various different methods for prediction and change detection 
    to detect gradual changes in the presence of non-trivial trends between CPs, while also keeping the false positive count low. 
    \item We instantiate \pandc{} with various modeling steps such as LSTM neural networks or ARIMA linear models, and evaluate it in a real-world case study in the realm of tribology, and show its advantages compared to several relevant reference methods.
    \item For this application, we developed a standardization to transform the data, which removes a linear trend from noisy data, thus increasing the visibility of other trends. After the transformation, any part of the data with only a linear trend is stationary. 
\end{itemize}

The paper sets the scene by defining the change point detection problem and reviewing related work, which we also use as reference approaches. We then state  three research questions, and subsequently introduce and formally define the adaptable and novel \pandc{} framework, which addresses these questions. A comparison of the framework with the reference methods concludes \Cref{sec:cpd}. 
\Cref{sec:exp_disc} contains the description of a real-world use case in tribology for online change point detection, and introduces the datasets we are working with. The datasets were pre-processed with a standardizing transformation, which we developed specifically for the given data. We also briefly discuss the implementation and the parameter choices for our experiments, for each tested method (\pandc{} and reference methods). 
In \Cref{sec:04_res_comp}, the results of the experiments with two different \pandc{} approaches, both with systematically tuned parameters, are compared to the results from optimally tuned reference methods in terms of false positive counts and out-of-control average run length. The sensitivity of the methods to different parameter settings is discussed also discussed. These results show that \pandc{} can deal with an online change point detection challenge arising from a real-world problem. Furthermore, it allows us to demonstrate how effectively \pandc{} solves that problem compared to other online change point detection methods. 
Finally, \Cref{05_Result_con}, the conclusion discusses the possibilities and limitations of \pandc{} and summarizes the results in the light of the three research questions initially proposed.

\section{Assisting CPD with Machine Learning (ML)}
\label{sec:cpd}

In this section, we will describe the proposed predict-and-compare (\pandc{}) framework in detail (\Cref{sec:pandc}), illustrate its operation on artificial data (\Cref{sec:artiData}), and subsequently compare it to several benchmark algorithms that we will use in the experiments described later in the paper. Before that, the relevant definitions, research questions (\Cref{sec:defs_resq}) and reference methods (\Cref{sec:ref_meth}) are presented and discussed.  In \Cref{sec:02_diffMeth} the \pandc{} framework is compared to the reference methods. 

\subsection{Gradual structural changes in the presence of trends}\label{sec:defs_resq}

We consider time series data in the form of real valued sequences $(X_n)_{n\in \R^\N}$, i.e., we look at finite sub-samples $(X_n^I)_{n\in I}$ with $I\subset \N$ of a sampled sequence of {\em random} real values $X_0, X_1, X_2, \dots X_n,\dots,$
where the index $n$ represents discrete time. More formally, we think of $X_n$ as the $n$-th image of a discrete-time stochastic process and $X_i^I$ as its restriction $X\upharpoonright I$ on the finite Index set $I$ (see Sect. 2.3). We assume non-stationarity of the probability measure associated with the random variable $X_n$. 
Changes in this distribution may be coming from {\em trends}, which means time-dependent changes of some of the distribution's moments in the form of a recognizable pattern. Alternatively, changes in {\em types of trends} are observed, meaning the pattern changes.  As an example, consider a Gaussian process with fixed variance but linearly (in time $n$) changing location parameter, which turns into a Gaussian process with fixed mean and increasing variance: This would indicate a change point between two parts of the time series in which the moments are changing by means of a constant, recognizable pattern.

Naturally, as only finite data samples are given and no details of the underlying distribution, it can only be {\em estimated} what is a trend and what is a change point. However, if changes in the data repeat as a pattern for specific scales of time (i.e., lengths of time-windows), then change points can be discriminated from these trends as singular, non-repeating changes. These repeating patterns can be learned and predicted.
Here, it is where machine learning (predictive modeling) helps in the formulation of the following definitions:\\[-0.4cm]

{\bf Definition 1}: A change in time $n$ of the distribution of a time series $X_n$ which can be learned and predicted from data observed in the past will be called {\bf trend}.\\[-0.3cm]

{\bf Definition 2}: Given a predictive model and a time series, a change in the distribution of a time series which cannot be learned and predicted by the predictive model is called a {\bf change point}.\\

In order to place the approach of goals A., B., and C. into a comprehensible framework, we formulate the following guiding questions:

{\bf Q1}: How can a change point in a time series be discriminated  from structural changes induced by trends?

{\bf Q2}: Is there a natural way to use predictive modeling to assist in recognizing change points in time series under the presence of trends?

{\bf Q3}: Does our proposed method compete well among state of the art online CPD techniques?\\

The answers, given in Section \ref{05_Result_con},  inform the reader how to use {\bf any given predictive model} capable of forecasting the progression of a time series from a window of observations to better discriminate between change point and artefact of a trend pattern. This summarizes the motivation of the theoretical concept of the approach: As long as trends are predictable, they can, using \pandc\, be discerned from truly unexpected changes in the parameters of the observations' distribution.

As a use case, we consider a tribological experiment. For the description of our method, the following terms are of importance in this context. Wear is a term relating to the interaction between and change of two surfaces, typically occurring as some relative motion between these surfaces leads to adhesion, abrasion, erosion or other kinds of mechanisms involving physical disturbances. It often happens that these disturbances are changing their intensity over time, dividing the life-cycle of the participating parts into three stages: The primary, or run-in regime, in which the asperities (microscopic high points) of the two surfaces are worn off to approach a state of equilibrium, characteristic of the secondary stage, in which a steady state with a constant rate of progression of the process is observed, and the tertiary stage, in which a progressive and divergent rate of change in the intensity of the disturbances prevails and usually leads to a destructive change of the involved machine parts.

Our data comes from a condition monitoring technique, based on radioactive isotopes.
It is necessary to distinguish between local statistical fluctuations, e.g., the radioactive decay process and real changes in the wear behavior in the measured signal (see \Cref{fig:02_origDataPlot}). Generally, the wear behavior can be differentiated into:\\

\begin{itemize}
\item Run-in wear (E), that is provoked by the adaption of a wear system to a change in loading conditions and characterized by a decreasing wear rate followed by a
\item steady-state (or constant) wear (K) that is easily characterized by a linear wear trend or a constant wear rate.
\item Divergent wear (A) is characterized by an increasing wear rate, which indicates or at least leads to the failure of the machine part. 
\end{itemize}

\subsection{Related Work}\label{sec:ref_meth}

We will compare the time until detection of \pandc{} as well as the number of false positives to various other state-of-the-art online CPD techniques \cite{burgEvaluationChangePoint2020}, particularly for the special case of heterogeneous data with non-trivial trends.  We pick four different representative methods from a wide variety of CPD approaches (a Bayesian approach, \cusum, \bfast, \ocd) to compare our approach to in terms of time until detection and number of false positives.

Our approach to online CPD in the presence of trends, called Predict and Compare (\pandc{}), will be formulated as a framework for using machine learning to assist in the differentiation between a trend and a change point. A predictive model (long short-term memory (\lstm{}) or auto-regressive integrated moving average (\arima)) makes a prognosis of the data in a (small) time window of the immediate past (the 'prediction window'). If there is a strong deviation between the real data on this time window and this prediction, a change point is found in it. For testing the deviation, we also use \cusum{}, thus naming the method \lstm{} \cusum, and \arima{} \cusum{}, respectively.

While statistical CPD techniques usually depend on a test statistic and connect the applied threshold with a significance level (\cite{amin_2017}: Def. 7), there is usually no explicit associated predictive model. On the other hand, 'anomaly detection' methods from the machine learning literature are often not cast in the test statistical context: See e.g. \cite{bukovsky2019}\cite{information15}. An exception to this is given in \cite{ma_rul_2021} by an approach on RUL-prediction with the aid of detecting capacity regeneration points. \pandc\; also combines the power of predictive modeling (for the recognition of possibly complicated trends as in-control non-stationary background signal) with the sequential test setting.  Other examples of non-linear, non-seasonal trends also occur in other heterogeneous data sets \cite{krause}\cite{foo}. Furthermore, \cite{amin_2017}, Def. 5 identifies sliding windows as a typical online CPD feature: CUSUM is recognized in the CPD literature \cite{SemiSupervised} as a classical detection method using sufficient difference between models on \mbox{sliding} windows. To the best of our knowledge, the combination of the CUSUM rule with the {\em assistance} of a predictive model in the form of \pandc{} has not yet been exploited for handling of heterogeneous data,  i.e. trends between CPs, which may change at a CP’s occurrence.

\subsubsection{Bayesian CPD}\label{subsec:02_Bay}
Bayesian methods have been successfully applied to online CPD \cite{pan_2005}\cite{AdamsMacKay}\cite{Fearnhead2007}\cite{diego_2020}\cite{malladi}\cite{pan_2022}). As shown recently, the method is also well suited to treat heterogeneous data \cite{lau}. However, they so far have been applied to time series data with stationarity between two CPs, excluding the case of trends between CPs.

The Bayesian approach to online change point detections rests on the idea that a predictive distribution at a particular time is used as a prior distribution for the location of the next change point. The prior is taken to depend on parameters of a model from survival analysis, given data known up to the current point in time.  More specifically, the prior is taken to be the survival function corresponding to the event of a change point occurring in the future ('survival' is the continuation of the time series {\em without} change point). The probability of a 'run' without change points of length $r_t$ to grow by one step is modeled by using the hazard function $H(t)$ expressing the rate of a failure (structural change) to occur during the interval of one discrete time step:
\begin{eqnarray} 
P(r_t>r_{t-1} \mid r_{t-1}) = 1 - H(t). 
\end{eqnarray}
Here, $H(t)=\int_{t-1}^t (s)ds$ with the hazard rate $h(t)$ being the usual logarithmic derivative of the survival function \cite{survival}. As we deal with heterogeneous trends, in which some parts are non-stationary even in the standardized form, without much further knowledge of the process, it is hard to accurately model the prior with a constant-in-time $H(t)$.
However, it has been shown in \cite{krause} that even heterogeneous data, as long as it is stationary locally (i.e., between the CPs), can be handled well with Bayesian CPD. In \cite{Fearnhead2007} and \cite{diego_2020}, this aspect is adopted from \cite{AdamsMacKay}, which is why we picked this method as a representative reference. We believe that using methods similar to the ones employed for our Predict and Compare algorithm, it can be extended to work in the non-stationary case, particularly if information about the intermediate time length's distribution used in the prior is available.

However, in one of the Bayesian approaches, the idea of a dynamic change of the learned ’regular data’ is considered. Namely, for the problem of how to prepare for the detection of change points by first learning the regular data in a ‘Phase I’ with insufficient information to estimate regular distribution parameters properly \cite{pan_2005}\cite{pan_2022}. These techniques to ’self-start’ an online CPD-detection without an initial learning process involve a sequential updating scheme, even in the case of complete prior ignorance. \pandc\; allows for such online updating of the current ’normality’ if the training process of the predictor is sufficiently fast. This is the case for linear predictive models such as ARIMA (in ARIMA-CUSUM, see Sect. 3.3.5 and Sect. 4.2 for the results). Window sizes of as small as $nh = 20$ consecutive data points are included in these studies. The advantage of our approach is that no assumption on the predictive distribution (input data’s distribution in \pandc’s case) has to be made (cf. \cite{pan_2023}, Sect. 3.1.3 - Scenario 3).

\subsubsection{Classical \cusum}
\label{subsec:02_class_CUSUM}

The classical \cusum{} rule \cite{pageContinuousInspectionSchemes1954a} in the standardized decision interval form \cite{woodallSTATISTICALDESIGNCUSUM1993,hawkinsCumulativeSumCharts1998} asks for the cumulative sum $S_j$, which is typically defined recursively:
\begin{eqnarray}
S_j\;=\;\max\left(0,\; S_{j-1} +X_j-\theta_j-k\right)
\label{eq:standardCusum}
\end{eqnarray}
where the 'target' $\theta_j$ is usually taken to be a running mean at time $j$, and  $k$ (the {\em allowance} making the detector less sensitive) is generally chosen half the step-size to be detected (see Eq. (2.3) in \cite{woodallSTATISTICALDESIGNCUSUM1993}), and specific to the so called decision interval form of \cusum{} (see \cite{hawkinsCumulativeSumCharts1998}, Chap. 1.9). The stopping rule is: Check in each time-step $j$ whether this sum exceeds the threshold $\lambda$, where \cusum{} locates the CP at the last (largest) time $j\in\{1, \dots, n\}$ at which $S_j=0$ (Note that usually the letter $h$ is used for the threshold; see \cite{woodallSTATISTICALDESIGNCUSUM1993} and \cite{hawkinsCumulativeSumCharts1998}, Chapter 2.1). To detect changes that have a negative deviation from $\theta_j$ some changes need to be made on Eq. \ref{eq:standardCusum}, exchanging the $max$ with a $min$ and adding $k$ instead of subtracting it. A change point is detected if the resulting value is smaller than $-\lambda$. Checking only for upward exceedances is useful to (increasing) wear related data and is applied in the Compare step of \pandc.

\subsubsection{Break detection For Additive Season and Trend: \bfast}
\label{02_bfast_exp}

Break detection for Additive Season and Trend (\bfast) is a change point detection introduced by \cite{verbesselt_detecting_2010}. In \cite{verbesselt_near_2012}, the method is used in an online setting. \bfast\; uses a Season-Trend model to model the data's seasonal aspect and trend. If there is no trend or seasonality in the data, that part of the model is omitted. This is the Season-Trend model from \cite{verbesselt_near_2012}:

\begin{equation}
\label{eq:02_bfast}  
y_t=\alpha_1+\alpha_2t+\sum^{k}_{j=1}\gamma_j\sin\left(\frac{2 \pi jt}{{\nu}}+\sigma_j\right)+\epsilon_t.       
\end{equation}

where $\alpha_1$ is the intercept, $\alpha_2$ the slope, $\gamma_1,\dots,\gamma_k$ the amplitudes, the season is $\delta_1,\dots,\delta_k$, $\epsilon_t$ is the error term at time $t$, $\nu$ is the number of observation per year and $k$ is the number of harmonic terms (this k is different to the $k$ in \cusum). The restriction to linear and seasonal trends is not given in \pandc. 

The Season-Trend model is calculated for two time periods to find change points in the data. One time period contains data from a stable history, and the other contains new data. The parameters of these models are compared with a statistical test ('Moving Sum' (MOSUM)) designed to detect changes in model parameters. In case of a change point, the MOSUM results will continuously deviate from zero.

\subsubsection{\bf Online Changepoint Detection: \ocd}
Chen et al. propose an online changepoint detection (\ocd) based on aggregated likelihood test statistics \cite{chen_OCD_2022}. A change is detected if one of two calculated statistics is above a threshold. The first is a likelihood ratio test statistic calculated for the last h data points per dimension. The test is between a known distribution and a simple alternative, for which the last h data points are used  ('tail sequence'- c.f.: Prediction Window). The most extreme of these likelihood ratios is then compared with a threshold.
 The second statistic is used to detect changes that are not concentrated in a single dimension. For dimension j, the partial sum of the tail sequence of the other dimensions is calculated (called off-diagonal statistic). Partial sum because the length of the used tail sequence is dependent on the result of the diagonal statistic for dimension j, which means not the whole tail sequence might be used. Again, this is compared with a threshold value to determine a possible change point occurrence.

\subsection{A new online ML-assisted CPD framework}
\label{sec:pandc}

The quality of CPD methods is measured by the 'average run length' $\textrm{ARL}_0$ (in-control) and the delay until detection $\textrm{ARL}_\Delta$ (the average run length 'out-of-control'). Ideally, the former should be as large as possible and the latter as small as can be. In statistical process control, the CUSUM test plays a dominant role, but there are other sequential test types (such as the Shewhart control chart and the EWMA sequential test). The simplicity and sensitivity of the CUSUM test makes it particularly interesting for generalizing it to handle more complicated underlying trends \cite{shar_2016}.

\begin{figure}[htbp] 
\centerline{\includegraphics[width=1.0\linewidth]{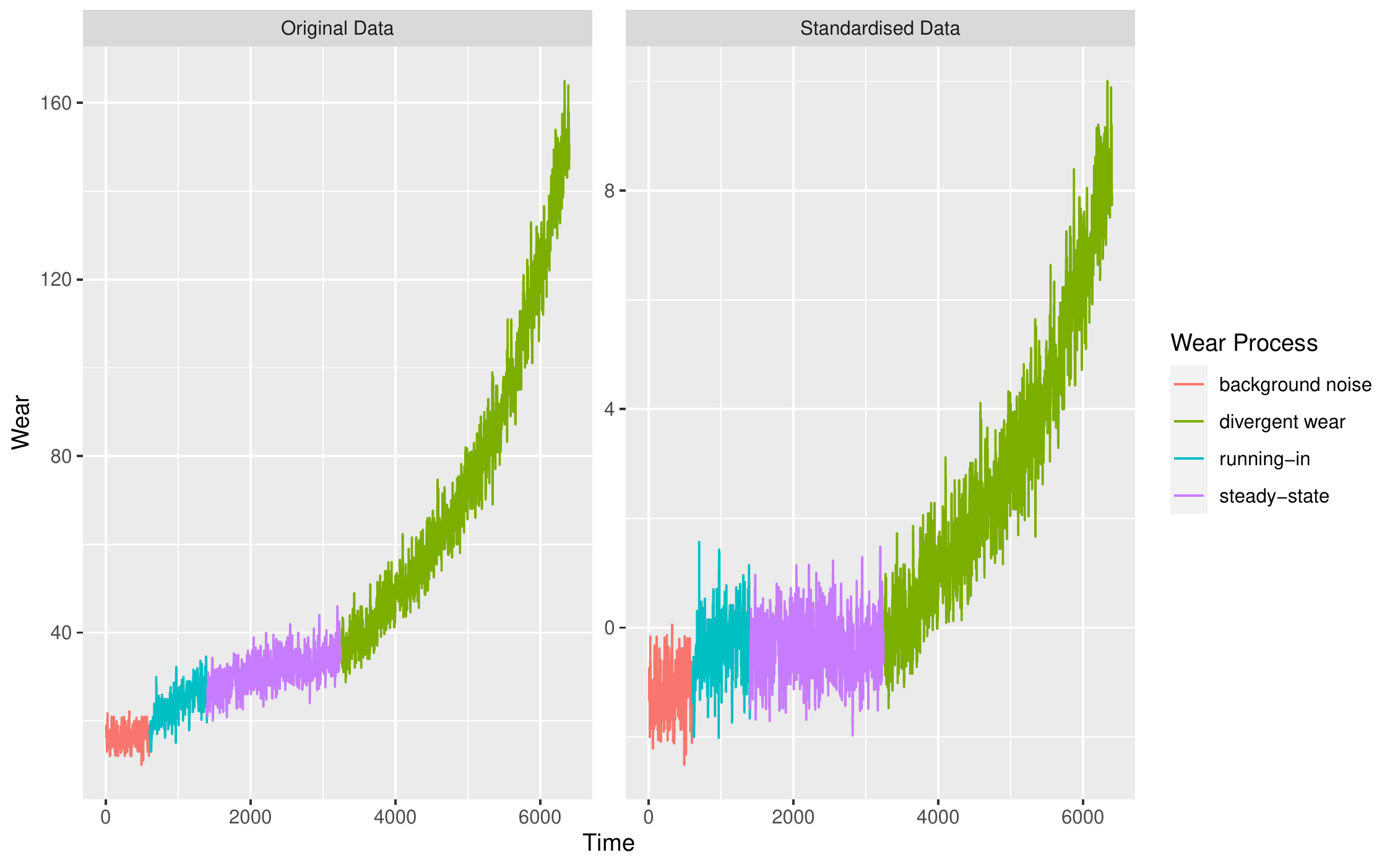}}
\caption{Heterogeneous time series data with regimes of different characteristic trends (left). See \Cref{subsec:03_dataExp}) for a detailed description of the tribological origin of the data. Also seen is a transformed version of the time series yielding a 'steady state' in one of the regimes (right), as described in \Cref{subsec:03_zScore}. Detecting change points into and out of stationarity will be seen to be simpler and add to the power of the detection method.}
\label{fig:02_origDataPlot}
\end{figure}

In Sect. \ref{subsec:03_dataExp}, we will give a more detailed description of the distribution belonging to a specific industrial (Tribological) application. These specifications, however, have nothing to do with the defined CPD-methodology other than it is particularly suited for this data type. \Cref{fig:02_origDataPlot} shows the typical form of input data with several time regimes forming characteristic types of trends. The points of time separating these regimes are the change points that are to be detected. This means that the change points of interest, here, are characteristic of changes in trends- as opposed to mere changes of constant parameters - reflecting property C of the type of CPD problem we are interested in (see Sect. \ref{sec:introduction}).

\begin{figure}[tbp] 
\centering
  \includegraphics[scale=0.65]{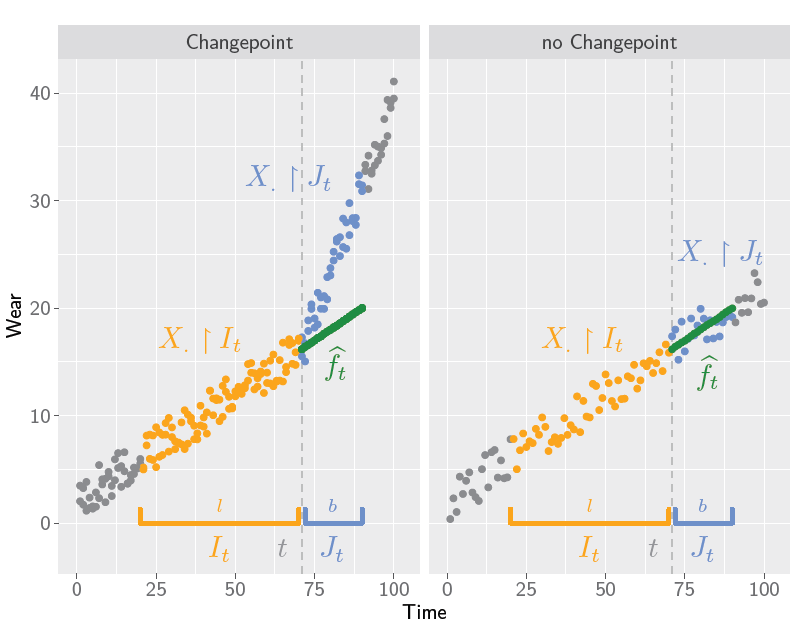}
\caption{An illustration of \pandc\ on data with a change point (left) and without a change point (right). The orange data points are used as input ($I_t$) for a predictive model $\widehat{f}_t$, whose predictions (green points) are then compared to the real data (blue points) on the prediction interval ($J_t$). The grey points are not used for the predictor $\widehat{f}_t$.
}
\label{fig:02_cpdMethod_helpPlot}
\end{figure}

The key idea of the proposed predict-and-compare framework is to apply a predictive model $\widehat{f}_t$ to input data from a time window $I_t$ of the immediate past, predicting a trend for future observations on a time window of the immediate future $J_t$. These predictions are then used to detect changepoints as significant deviations in the observed data from the predicted trend. This is illustrated in \Cref{fig:02_cpdMethod_helpPlot}, where the left part shows a case where a changepoint should be detected, and the right part illustrates a case without a change point. The process is described in detail as follows:\\

(1.) {\bf X and Y of the Predictive Model}: Let $l$ and $b$ be the two positive integers which refer to the respective sizes of the {\em input window} $I_t:=\{t-l+1, \dots, t\}$, and {\em prediction window} $J_t:=\{t+1, \dots, t+b\}$, located around the discrete time value $t\ge l$. For a given sequence of real numbers $X\in\R^\N$ (the signal),  the restrictions of this function on the positive integers to $I_t$ and $J_t$ will be denoted by $X\upharpoonright I_t$, and $X\upharpoonright J_t$, respectively. Two such vectors of size $l$, and $b$ are respective elements of the spaces of functions $\mathcal{X}:=\{X:I_t\to \R\}\cong\R^l$, and $\mathcal{Y}:=\{Y:J_t\to R\}\cong\R^b$. \\

(2.) {\bf Hopping Windows}: We partition the positive integers greater than $l$ into disjoint intervals of width $b$, namely $\N=\uplus_{t \in T} J_{t}$  where $T=l+b\cdot\N$. We consider, for each $t\in T$, apart from $J_t$, the input window $I_t$ of width $l$. Moreover, for $J_t$ we consider an increasing sequence of growing sub-intervals $J_t^j=\{t+1, \dots, t+j\}$,  of final maximal size $b$. The increasing times $s=t+j$ correspond to passing through the current moments in time within $J_t$.\\

(3.) {\bf Model training (fitting)}: We define a {\em learning method} $\mathcal{A}$ to be a map from a \emph{training data} set $\mathcal{T}:=\{\langle X^{(i)}, Y^{(i)} \rangle\}_{i=1}^n\subset \mathcal{X}\times\mathcal{Y}$ of size $n$ to a \emph{predictive model}, i.e., a function $\widehat{f}_t: J_t\to \mathcal{Y}$ 
that allows to approximate $Y\in \mathcal{Y}$ with $\widehat{f}_t(X)$, also for new and previously unseen data points $X \in \mathcal{X}$.  We perform the training of a learning method $\mathcal{A}$ once and use the training data set $\mathcal{T}$, where the vectors $X^{(i)}$ and $Y^{(i)}$ are taken from a different time series sample representative of the regular (change point free) part of $X$. \\

(4.) {\bf Predict and Compare}: So, for the current time $s$, let the largest multiple of $b$ plus a single input window size $l$ smaller than $s$ be given by $t=l+b\cdot m$ for a suitable $m\in\N$. Then, the vector $X\in\mathcal{X}$ of values from the current input window is inserted into the predictive model $\widehat{f}_t:\mathcal{X}\to\mathcal{Y}$. The first $j$ values of the associated prediction $\widehat{f}_t(X)$ can then be compared to the first $j$ values of the vector of real observed data $Y$ on $J_t$. At each of these times $s=t+j$, a sequential test is run and either a detection is found or the current time $s$ is incremented by one. \\

(5.) {\bf CUSUM in Compare step}: The test we use in the Compare step is the \cusum{} test  (\ref{eq:standardCusum}). It acts as a map $C:\mathcal{Y}\times\mathcal{Y}\to\{\textrm{CP, No CP}\}$ to the possible outcomes of the test. In the following recursive expression, the \cusum{} rule for {\em upward} changes is defined using the $\widehat{f}(j)$, the
$j$-th element of the predicted time-series on $J_t$ as the target $\theta_j$:
\begin{equation}\label{eq:indicator}
    S_j\;=\;\max(0, S_{j-1}+X_j-\widehat{f}_t(j)-k\,),  
\end{equation}
where, as in Step (2.), $t=\max\{r=b\cdot m + l\,\vert \,r\,<\,s\;,\; m \in \N \}$.  In this way, \cusum{} is 'assisted' by the predictor $\widehat{f}_t$.\\[-0.3cm]

{\bf Remarks}:  In step (3.), we choose the training set and learning method according to the trends we wish to consider as 'regular data'. It is the deviations of them that define a change point. This answers question {\bf Q1}.

{{\bf Q2} is answered by (4.), in general, and (5.) in the case of a particular choice of the sequential comparison test: The trend is predicted by $\widehat{f}_t$ while the CP is detected by the help of the prediction-assisted CUSUM-rule.  We choose  $C: \mathcal{Y}\times\mathcal{Y}\to\{\textrm{CP}, \textrm{No CP}\}$  by letting the indicator in (\ref{eq:indicator}) decide  whether there are deviations between prediction $\widehat{f}_t$ and data $X\upharpoonright J_t$. The CUSUM test makes the whole detection method be {\em online} (see discussion following the definition of the algorithm), and also because it provides a more exact location of the CP than the end of the prediction interval at the time of detection. \\

In addition to the classifier $C$, it is useful to consider a separate locator $L:\mathcal{Y}\times\mathcal{Y}\to J_t$ deciding about the time of occurrence of the CP. In the simplest case, it might coincide with the time of detection. In more sophisticated cases, the CP is placed somewhere in $J_t$. E.g., CUSUM uses the last time $s=t+j$ at which the indicator $S_j$ is zero \cite{bassevilleFAULTISOLATIONDIAGNOSIS2002}.\\

{\bf Definition 3}: {\em Predict and Compare Detector} -- A Predict-and-Compare detector (P\&C) is a tuple $\langle \mathcal{A}, \mathcal{T}, l, b, C, L\rangle$ in which $\mathcal{A}:\mathcal{T}\to(\mathcal{X}\to \mathcal{Y}$) is a learning method defined on the training set $\mathcal{T}$ with values in the set of predictors mapping size-$l$ inputs from $\mathcal{X}$ to size-$b$ predictions in $\mathcal{Y}$, a classifier $C:\mathcal{Y}^2\to \{\textrm{CP}, \textrm{No\,CP}\}$ deciding if a sub-sample of $X$  of size $b$ is significantly deviating from a target sequence of the same size, and a locator $L:\mathcal{Y}^2\to J_t$ naming the estimated point in time of the CP.

\begin{algorithm}
\caption{\pandc{}}
\begin{algorithmic}[1]
    \Procedure{\pandc}{$\mathcal{T}, \mathcal{A}, X, l, b, C, L$}
    \State $\widehat{f}_t=\mathcal{A}(\mathcal{T})$, the predictor
    \For{$t \in l+b\cdot\N $}
        \State $I_t = \{t-l+1, \dots, t\}$, the current input window
        \State $J_t = \{t+1, \dots, t+b\}$,  the current prediction window
        \State $\widehat{Y} = \widehat{f}_t(X\upharpoonright I_t)$, the current prediction in $\mathcal{Y}$
        \For{$j \in \{1,\dots, b\}$}
        \State $J_t^j=\{t+1, \dots, t+j\}$
        \State Let $\widehat{Y}\upharpoonright J_t^j$ be the vector of the first $j$ elements of $\widehat{Y}$.
        \If{$ C(X\upharpoonright J_t^j, \widehat{Y}\upharpoonright J_t^j)==\textrm{No\,CP}$}
            assume no CP in $J_t^j$
        \Else{
            CP in $J_t^j$} 
            \State Record $j$ as the point in time of the CP detection.
            \State Record $c_j=L(X\upharpoonright J_t^j, \widehat{Y}\upharpoonright J_t^j)$, the CP location.
        \EndIf
        \EndFor 
    \EndFor
    \EndProcedure
\end{algorithmic}
\end{algorithm}
\vspace{0.3cm}
Note that even though the algorithm is 'windowed' into a discrete, non-overlapping, exhausting set of sub-samples, it can still be used online because the comparison method $C(\cdot, \cdot)$ is online. \\

A true online detector can be applied at each accessible moment in time, i.e. for every moment at which the information belonging to the current time step becomes available. Sliding windows (such as in \bfast, cf. \Cref{02_bfast_exp}) or a cascade of fixed windows partitioning the time axis (as in our proposed detection algorithm) may be part of the approach without impairing the online property. While the current point in time makes the sliding window move continuously along, the cascading window approach lets it wander through each consecutive interval, in which a parameter adaption is made. Nonetheless, the online property is completely retained, it is just a step-wise parameter-adjustment at the interfaces between two consecutive time windows. \\

Also, instead of changes in the location parameter, changes in the local dispersion of the signal can be controlled by CUSUM, e.g., by using the squared data $X_t^2$. There is classical work \cite{bissell2} on asymptotic optimality of such control chart problems from sequential modeling and specific CUSUM rules for CPs in the variance of a signal \cite{variance}. All of these can be realized as CP-types in the \pandc{} framework by choosing the corresponding CUSUM test in the Compare step.

\subsection{Analysis of artificial data}
\label{sec:artiData}

Before applying the \pandc\ method to experimental data, we first use artificial data with different types of change points to allow for a broad orientation of what to expect. As described in Sect. \ref{sec:defs_resq}, the process of physical wear in tribological systems undergoes several stages. As described in more detail in Sect. \ref{subsec:03_dataExp} the condition of a blue is 
monitored: Particles are coming loose from the pair of surfaces and initial abrasive wear due to physical contact in the run-in regime is followed by hydrodynamic effects, including concatenation of the hydrodynamic contact-less lubrication. The transition into the divergent regime happens as particles amass and progressively damage the tribological pair given by bearing and shaft to the effect of there not being enough space for contact-less operation. The condition-monitoring involves counting radioactive decays proportional to the particles loosened by the wear process \cite{jech_radionuclide_2018}.

To get an idea of the power of P\&C on such data, we generate time series samples with a simple model representing a change point from the run-in to the steady-state phase and another change point from the steady-state to the divergent wear regime.

For the sake of simplicity and to compare different forms of change points under different noise levels, we model the rate of wear by
\begin{eqnarray}
f(t) = a\cdot\lambda\cdot e^{-\lambda t}\;+\;c\;+\;d\cdot {\bf I}_{{}_{[t_2, \infty)}}(t)\cdot\left(t-t_2\right), 
\end{eqnarray}
where $\lambda, a, c, d \in {\bf R^+}$ \cite{AC1}. It is seen that after the decay of the exponential run-in process of amplitude $a$ to a (arbitrary, but fixed) fraction of one (at some time $t_1$) with rate $\lambda$  a phase of purely steady-state ('linear') wear settles in and the rate progresses with constant magnitude $c$. The second change point occurs later (at $t_2$) when divergent ('quadratic') cumulative wear becomes the dominant part.

To use the simplest renewal process with this time-dependent rate, we employ the time-dependent Poisson process \cite{Pstandard} with intensity function, which is given by $f(t)$ (usually called $\lambda(t)$, see \Cref{subsec:03_zScore}) to simulate data shown in \Cref{fig:02_genDataRes}. The lower part of this visualizes the aggregation (by mean) of our two quality measures (\Cref{sec:041_qualitMeasures}) Fpc and ArlP for all data samples over different tuning parameter sets for Predict and Compare. A first impression shows that the data sample classes belonging to the three signal-to-noise ratios group into three clusters 1,2,3, in which group 1 has the lowest and group 3 has the highest signal-to-noise ratio. \\

Overall, the results from the aggregation of the two considered CPs (lower line of diagrams of \Cref{fig:02_genDataRes}) show that the signal-to-noise ratio has a significant influence on the result regardless of the parameter settings for P\&C: From the aggregated results, it can be observed that with increasing noise level \\

\begin{itemize}
    \item[(a)] for the divergent CP (into regime A), the Fpc is increasing, and the ArlP is first increasing and then decreasing while
    \item[(b)] for the CP from the run-in into the linear regime (into regime K), the Fpc is first decreasing and then increasing, while the ArlP is decreasing. \\
\end{itemize}

Observation (a) correlates with the intuitive notion that an increase in noise will increase the number of false positives. However, the time until detection being the largest for the {\em intermediate} noise level is surprising.  Similarly, observation (b) seems intuitively clear the final increase in Fpc goes. The decrease of Fpc and ArlP is unexpected.

This shows that a change in data may, on the one hand, reduce the power of the \pandc{} detector. However, it may also improve the power of the trend predictor $\widehat{f}_t$, which leads to improvements in terms of Fpc and ArlP. In \Cref{04_Result_exp}, we investigate this systematically for a series of industrial data sets from tribological experiments.

\begin{figure}
    \centering
    \includegraphics[width=1.0\linewidth]{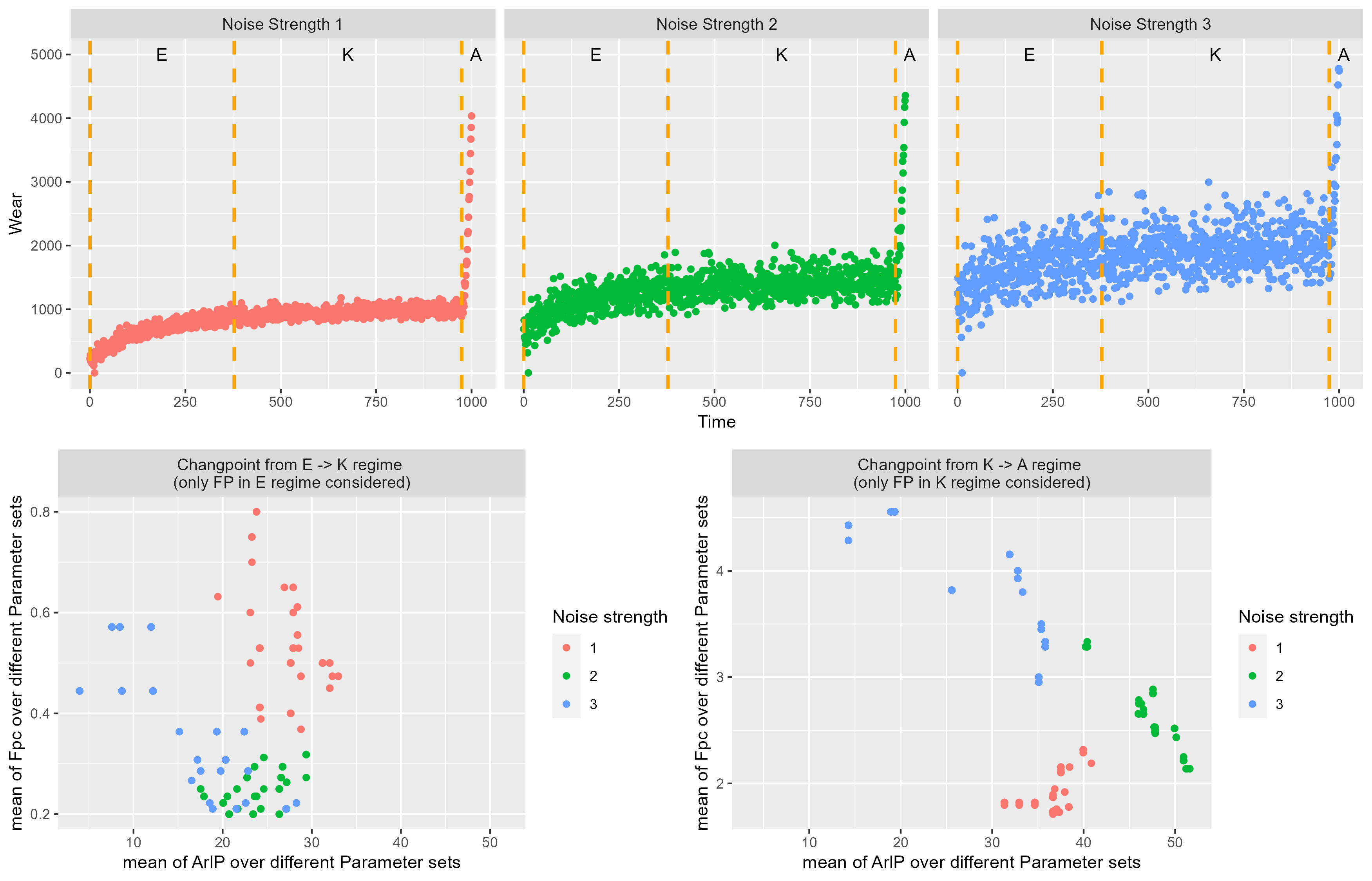}
    \caption{The top three plots show artificial data samples used to test the Predict and Compare method. The second and third vertical dashed orange lines symbolize the CPs into regime K and regime A, respectively. Bottom: The plot shows the results of \pandc{} with \lstm{} as the learning method applied to these two change points.  The colors signify the respective strength of the signal-to-noise ratio from the diagrams above. Each point represents the averaged results of \pandc{} over different parameter sets for one data sample. The x-axis is the difference between the labeled and the found change point. The false positive count (Fpc) is shown on the y-axis. Both averages are aggregated by using the mean. }
    \label{fig:02_genDataRes}
\end{figure}

\subsection{Relation of P\&C to other approaches}
\label{sec:02_diffMeth}

In this section, we compare the reference methods (\cusum{}, \bfast{}, OCD and Bayesian CPD) to \pandc{} and show the similarities and differences between them.

\subsubsection{Classical \cusum} 
The difference between classical \cusum{} and Predict and Compare is that the target (or 'quality number', \cite{pageContinuousInspectionSchemes1954a}) $\theta_j$ is replaced by the prediction $\widehat{t}_t(j)$. These predictions are subject to the result of the trained model $\widehat{f}_t$ and its input given by the data from the input window $I_t$. The predictions $\widehat{t}_t(j)$ are valid throughout a single prediction window $J_t$, and updated, as soon as the current time $j$ enters the next prediction window.

\subsubsection{Break detection For Additive Season and Trend (\bfast)}
In contrast to \pandc, where the prediction $\widehat{t}_t(j)$ is compared to the real data points $J_t$, \bfast{} compares the parameters of the two models, using a 'Moving Sum' (MOSUM)\cite{bfastMosum_zeileis_2005}. MOSUM is a statistical test designed to detect changes in model parameters. 
If MOSUM detects a parameter difference, the historical time period ($I_t$) is different from the monitored time period($J_t$), and a change point is detected. On the other hand, there is no change point if no difference is found as both periods are the same. 
Choosing the right size for the historical time period is vital for a successful comparison. \bfast{} offers an automated method to find a good value for $l$. Expert knowledge about the data can also be used to define $l$. 

\bfast{} uses a historical and a monitoring time period similar to Predict and Compare. Therefore, it is interesting to compare those two methods. \\

\subsubsection{\bf Online Changepoint Detection (\ocd)}
The \ocd{} aggregates a value gained from the data and compares it to a threshold to find change points, similar to \pandc. For \pandc{}, this value is the sum of differences between the real data point and a predicted data point, which indicates a change point if it crosses a threshold. \ocd{} calculates the diagonal statistic and off-diagonal statistic for the threshold comparison.

\subsubsection{\bf Bayesian CPD}

The Bayesian approach to online change point detection taken in \cite{AdamsMacKay} potentially using more than one latent state employs updating the posterior distribution of the run-length $r_t$ during every time step using the hazard function $H(t)$. If the Hazard function is constant, the run-length distribution becomes geometric and independent of the observed data (see \cite{diego_2020}, Sect. 2.2). Similarly, if $H(t)$ is bounded from below by a positive constant $c$, then the distribution of the residual time $l_t$ before a CP occurs is also of the form $c(1-c)^{l_t}$. This situation is given if the data before the occurring CP is stationary. This occurs in our use case after the standardization transformation has been applied (see \Cref{fig:Example_Z}, the stretches before the green (divergent) regime).

Bayesian methods such as that of Adams and MacKay are fully online, helping the time until detection become small. The predictive distribution of future values of the time series values given the past observations has to be calculated based on the posterior distribution of the run length. For very small Hazard rates, however, it becomes hard to determine this with sufficiently high predictive accuracy. Experiments with very low rates of occurring CPs (such as changes between regimes of tribological wear and non-trivial trends) are thus difficult to approach with this predictive filtering approach.

\section{Experiment}
\label{sec:exp_disc}

In this section, we will describe the details of the experiments conducted for this paper. In \Cref{subsec:03_dataExp} the source of the data and its characteristics will be described. Then a transformation developed to preprocess our specific data will be introduced in \Cref{subsec:03_zScore}. In \Cref{subsec:03_parameter} and \Cref{subsec:03_impl} the parameter selection and implementation details for each method are explained.

\subsection{Data for Experiments}
\label{subsec:03_dataExp}

For the evaluation of the CPD methods/strategies, data sets of real experiments have been deployed within this paper. Within these experiments, the performance of machine parts is examined through a bench test (see \Cref{fig:triboBenchTest}). The wear of the specific and critical machine part is continuously monitored via an in-situ technique based on radioactive isotopes \cite{jech_radionuclide_2018}.

\begin{figure}[htbp]
    \centering
  \includegraphics[width=0.5\linewidth]{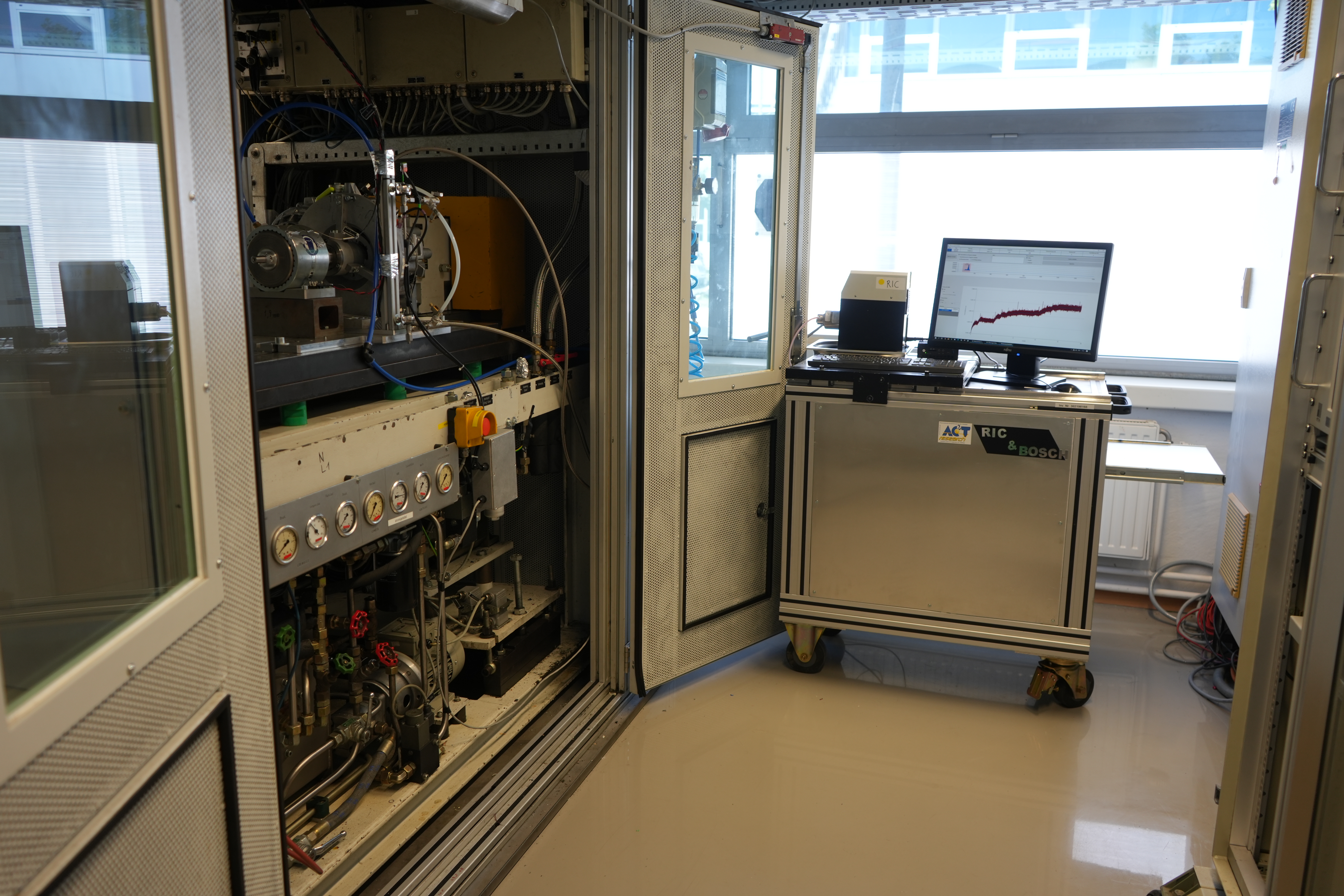}
    \caption{his picture shows a bench test setup. On the left side, one can see the test bench. Below is the supply hydraulics and in the upper area, a wear test has been set up. From the wear test, two hoses run horizontally to the right and continue downwards. The lower hose transports the lubricant from the wear test to the RIC, located to the right, and the upper hose returns the lubricant to the wear test. On the right side, one can see the computer used to monitor the test. (Photograph taken by Dr. M. Jech, publication granted with courtesy of the Austrian Competence Centre for Tribology, AC$^2$T research GmbH) }
    \label{fig:triboBenchTest}
\end{figure}

For this wear measurement, the critical machine part is labeled with radioactive isotopes via thin layer activation \cite{brisset_radiotracer_2020}. Due to the frictional contact, wear particles are generated and transported by the lubricant circuit to a gamma ray detector. Through the measurement of the activity of the wear particles in the lubricant and the combination with the knowledge about the isotope depth distribution in the labelled machine part, the amount of wear can be calculated. 

There is a shift of time, when a wear particle is generated until it reaches the detector and is detected. This time shift is in the range of a few minutes and consequently much faster than changes in the wear behavior. Nevertheless, this time shift must be considered for the comparison of different measurement methods (e.g., a rise of temperature) for in-time recognition of change points in machine performance.

The applicability of AI methods for CPD is dependent on the data provided by the measurement and thus the specific characteristics of the measurement method must be considered. For the applied continuously monitoring wear measurement, the statistics of radioactive decay must be regarded as part of the signal scattering/noise. Furthermore, wear particles are sometimes not distributed homogeneously in the whole lubricant circuit and so fluctuations of wear particle concentration may occur in the detector volume, which is a certain fraction of the whole circuit. These effects will be considered by specific standardization of the Poisson process (see therefore \Cref{subsec:03_zScore}).

The fast detection of the change points– for example when divergent wear starts – is crucial for wear testing but also for maintenance of machine parts in the production process. If the divergent wear change point can be detected timely and with high accuracy, the origin and cause of the wear can be investigated (especially interesting when testing prototype parts) and expansion of the damage can be limited in the production processes. In this sense, timely is referred to as being faster than the occurrence of the final damage but also being faster than other detection methods which are currently used. High accuracy refers to issuing a warning (signal) when divergent wear occurs. Divergent wear should not be overlooked by the CPD, but warnings without actual cause should also be avoided.

In total, four different wear phases can appear in the five uni-variate time series datasets used in this paper (\Cref{fig:all_data}). One phase is called experiment paused, which refers to times when no new particle from the experiment reaches the detector. The beginning of each dataset represents the stationary background noise detected by the detector. Another phase is the non-stationary running-in or run-in wear phase, where the increase in the wear volume starts high and reduces over time. This behavior is only visible in dataset 1 and 3. In the other three datasets, this is hidden by the noise. In the constant or steady-state wear phase, the increase is linear. In the fourth phase, the divergent wear, the wear volume increase is more than linear. The transition between two of the four phases is a change point.

\begin{figure}[htbp]
    \centering
  \includegraphics[width=1\linewidth]{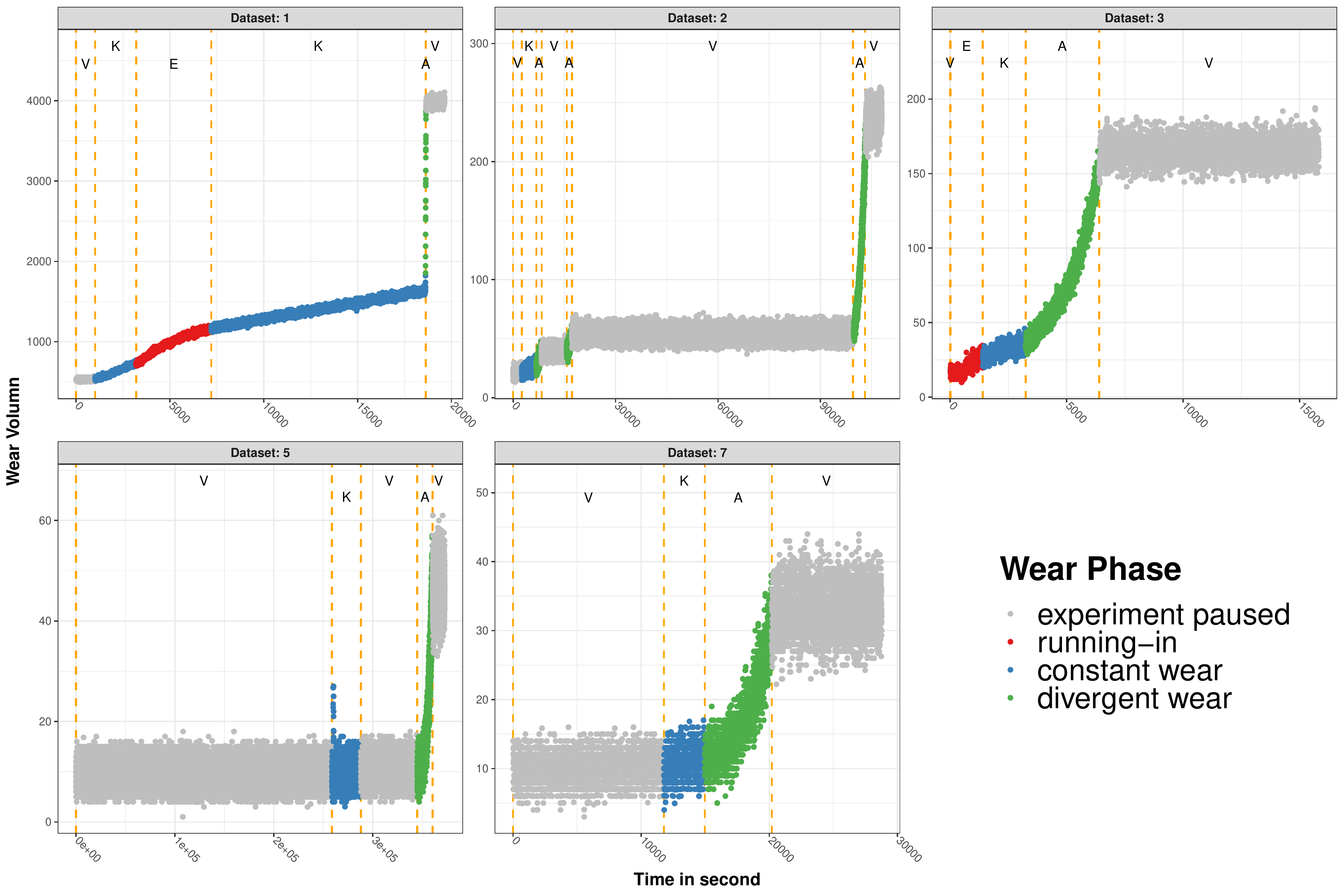}
    \caption{The five different data sets with divergent wear are used for evaluating the different change point methods.}
    \label{fig:all_data}
\end{figure}

\subsection{A standardizing transformation }
\label{subsec:03_zScore}
\pandc{} is evaluated with tribological data (\Cref{fig:all_data}), which due to the nature of the tribological experiments (see \Cref{subsec:03_dataExp}), are noisy. A transformation is applied to increase the detectability of the wear phase changes in the noisy data. For this transformation, we look at the cumulative sum of a noisy, and itself increasing function $f(t), t\in {\bf R}_+$ (representing the rate of wear over time $t$) with the independent but increasing in its standard deviation noise $W_t$. This leads us to a model for the wear per unit time $X_t$: 
\begin{eqnarray}\label{eq:Xrep}
X_t \;\;\;=\;\;\;f(t)\;\;+\;\;W_t\;\;+\;\;I_t.
\end{eqnarray}
$W_t$ and $I_t$, and so also $X_t$ are $t$-indexed random processes (i.e. for each $t\ge 0$ they are functions on a suitable probability space), and $f:[0, \infty)\to[0, \infty)$ is 'deterministic'. The function $t\mapsto I_t$ is either zero, or, represents a non-vanishing anomaly describing the structural change following the change point. 

We assume that within the regime of steady-state wear, there is arithmetic growth of the wear rate, namely $f(t)=b\cdot t^\nu$, where $b, \nu>0$. Given any estimator of $b$ and $\nu$ trained on data $\mathcal{T}:=\{X_i\}_i^n$ called $\widehat{b}$ and $\widehat{\nu}$, respectively:\\

\begin{algorithmic}[1]
    \Procedure{Standardization}{$X_t;\, \mathcal{T}, t_0, t$}
    \State Determine $\widehat{\nu}$ using $\mathcal{T}$.
    \State Determine $\widehat{b}$ using $\mathcal{T}$.
    \State Define $\widehat{\lambda}(t):= \widehat{b}\cdot t^{\widehat{\nu}}$
    \State Define $\widehat{Z}(t):=\frac{X_t\;\;-\;\;\widehat{\lambda}(t)}{\sqrt{\widehat{\lambda}(t)}}$
    \State \Return $\widehat{Z}(t)$
    \EndProcedure
\end{algorithmic}
\vspace{0.3cm}

The formula for $\widehat{Z}(t)$ is chosen with the idea that they are realizations of an inhomogeneous Poisson Process with an intensity function $\lambda(t)$, which represents the local mean {\em and} variance. Therefore, the standard deviation estimate results from the square-root of $\widehat{\lambda}$ in the standardization.

Now, estimating $b$ and $\nu$ follows the following argument (which we used in the calculation for the experiment in \Cref{04_Result_exp}): Assume that there is an estimator $\widehat{\Lambda}$ of $\int_{t_0}^t\lambda(s) ds$. Then, it is clear that if $t$ becomes large, under our assumption about $f(t)$, (\ref{eq:Xrep}), and the fact that because of  ${\bf E}[W_t] = 0$, we have $f(t) = d/dt{\bf E}[\int_{t_0}^t X_s ds] = \lambda(t)$, $ \frac{\log \Lambda(t_0, t)}{\log t} \to \nu + 1.$
Therefore, we choose  the estimation procedure 
\begin{algorithmic}[1]
    \Procedure{Estimate-Trend-Parameters}{$\mathcal{T}, t_0, t$}
    \State Compute $\widehat{\Lambda}(t_0, t):=\sum\limits_{j=t_0+1}^tX_j$ with the data in $\mathcal{T}$.
    \State $\widehat{\nu}_t:=\frac{ln(\widehat{\Lambda}(t_0, t))}{ln(t)}\;-\;1$
    \State $U_t:=t^{\widehat{\nu}}$
    \State Call $\widehat{b}$ the estimated coefficient $b$ of the linear regression model $$X_t=b\cdot U_t + \varepsilon_t.$$ \hspace{0.7cm}and use the data in $\mathcal{T}$ for fitting this model.  
    \State \Return $\langle \widehat{\nu}, \widehat{b}\rangle$
    \EndProcedure
\end{algorithmic}

The standardization also works {\em online} in the sense that up to every current point in time $t$, the available data $\{X_s\}_{s\in[t_0,t]}$ is input to {\em Estimate-Trend-Parameters}($\cdot, t_0, t$). \Cref{fig:Example_Z} shows the effect of this transformation on the run-in, the intermediate steady-state wear, and the terminating divergent wear regime. As in the intermediate regime, there is a constant arithmetic form of growth of the underlying trend function $f(t)$, the standardized data is (close to) a stationary signal, most of that regime. It is seen that there is a run-in phase at the beginning of the intermediate phase before stationarity 'kicks in'. Then, however, it is seen that the growing dispersion of the original data is transformed into a sequence with locally constant standard deviation. In spite of this standardization of the steady-state regime, the change point into the divergent regime, in the end, is still clearly marked by a visible change point. 

The key feature of the standardization may be considered to be line 3 of the Estimate-Trend-Parameters procedure. Here, the estimator $\widehat{\nu}_t$ is chosen for functions of arithmetic growth, i.e., of the type shown in line 4 of the Standardization procedure. This is the type of growth rate expected from the steady state regime in the original (non-standardized) data set.

A characteristic shape of the transformed sequence is visible in the first and third of the five used data sets shown in \Cref{fig:all_data} and \Cref{fig:allData_Z}. The data of the red run-in phase is transformed from an increasing shape into a curved non-monotonic shape, followed by the steady-state regime's eventually stationary stretch. The green stretches represent the divergent wear behavior, which is also clearly distinguishable from the steady-state wear regime in the standardized form. 

We conclude that the standardization yields a transformation into the eventually stationary form for the steady-state regime. At the same time, the change points into and out of this interval are still clearly detectable by the change point detector. This will become more clear in the discussion of the results (see \Cref{subsec:04_param}): Our detector defined by \pandc{} sees the run-in to steady state change point clearer than the reference methods. At the same time, the strong change point at the beginning of the divergent wear regime typically (among the five different experimental wear data sets considered) remains most clearly detectable, even though the standardized data is used.
\vspace{-0.5cm}

\begin{figure}[htbp]
    \centering
 \includegraphics[width=0.8\linewidth]{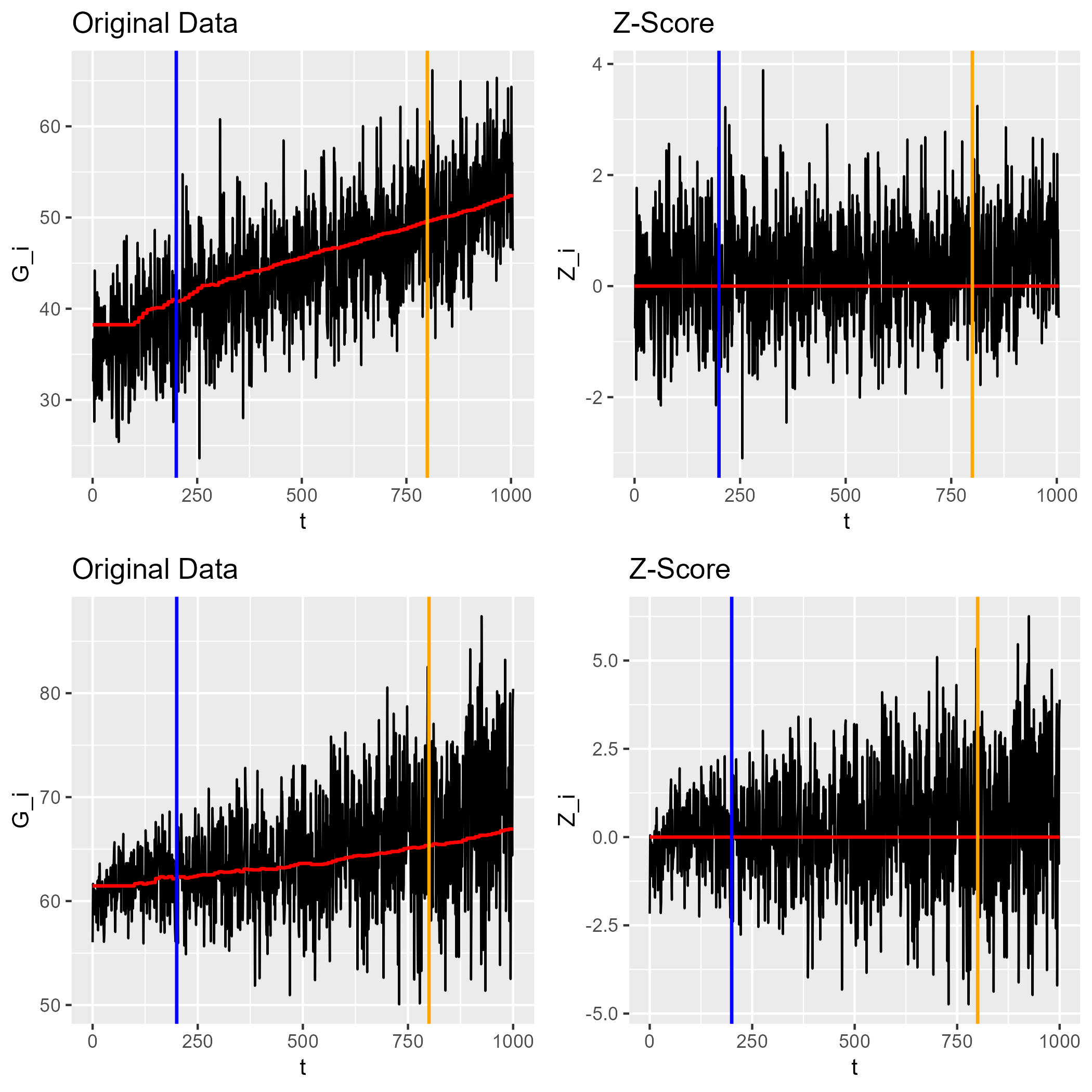}
    \caption{Examples of the standardization (right) of typical wear curves (left) under strong noise with run-in, steady-state-, and divergent wear. Both simulations in the left column show simulated inhomogeneous Poisson Processes (the red line being the intensity function). The standard deviation increases proportional to the square-root of time. The simulation in the upper row has the starting point moved to negative values, which is why the increase is not as clearly visible as for the simulation in the lower line. However, it is typical for the real experimental data, so the standardization effect is demonstrated here.}
    \label{fig:Example_Z}
\end{figure}

\begin{figure}[htbp]
    \centering
 \includegraphics[width=1\linewidth]{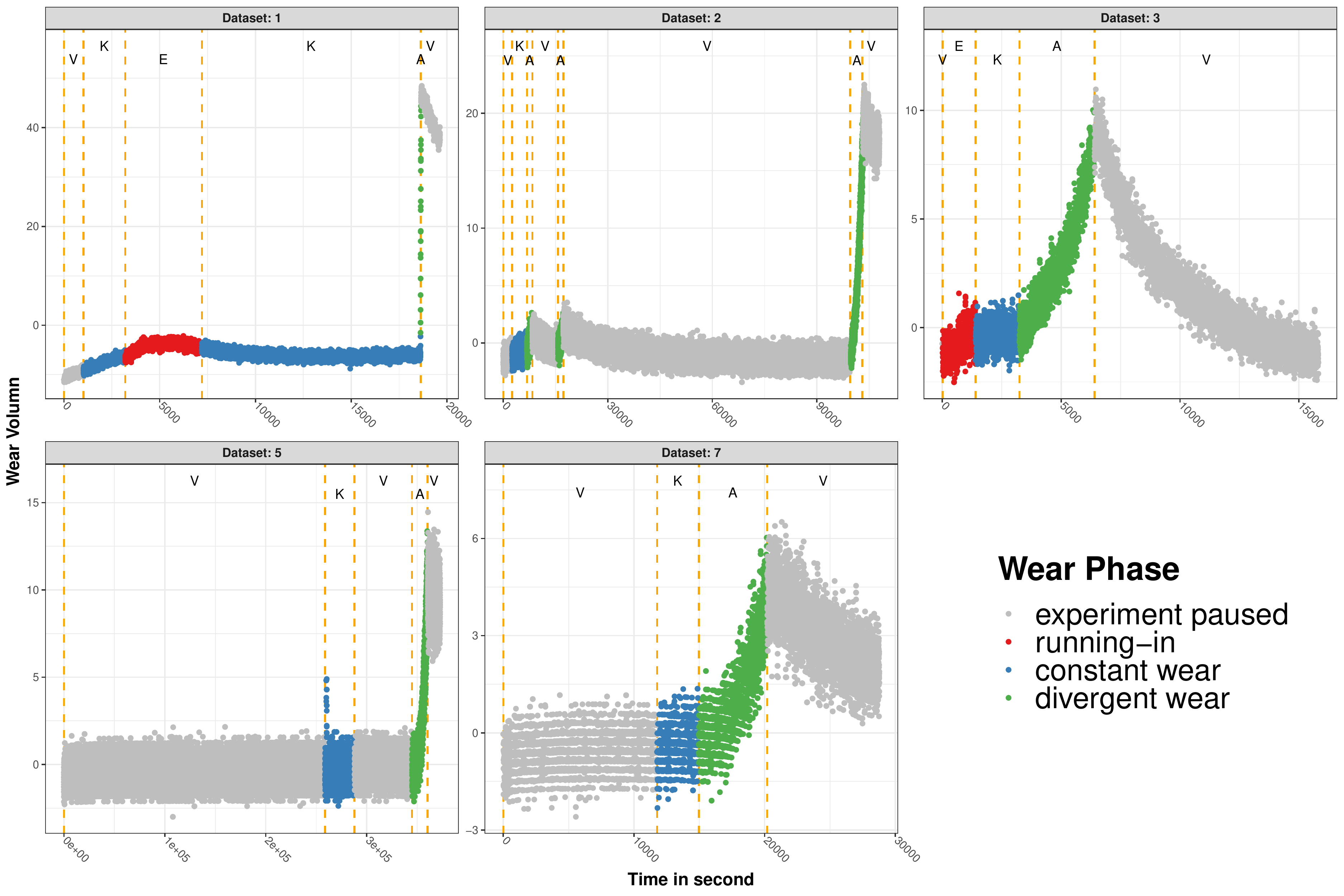}
    \caption{The same 5 data-sets as in \Cref{fig:all_data} standardized with our standardization method \Cref{subsec:03_zScore}. It is visible that the regimes of linear increase of wear yield a stationary z-score, after some 'run-in' period. Note that the standardization is calculated at every instant in time only with values available up to that instant, as is required by a full online detector.}
    \label{fig:allData_Z}
\end{figure}

\subsection{Comparative Experiments}
\label{subsec:03_impl}

All experiments were conducted with R. 
One experiment can be defined as using a method with a certain parameter set on one dataset. The change points found by an experiment were saved and later used to determine the effectiveness of each method by calculating two criteria, see \Cref{sec:04_res_comp}. 

Five different tribological datasets (\Cref{fig:all_data}) are used to evaluate and compare the change points methods. All these datasets include a divergent wear part. 
As part of the preprocessing, these datasets are standardized using the method described in \Cref{subsec:03_zScore} (datasets after the standardization \Cref{fig:allData_Z}). The methods used in this part of the paper are \bfast, Bayesian, \cusum, \arima{} used with \cusum{} and \lstm{} used with \cusum. 
Additionally, a simple 'baseline' is created by sampling random points from the datasets and treating them as change points. Recognizing the inherent independence and uniformity of the sampling procedure, it is apparent that conditioning the sampling based on a predetermined quantity of false positive finds can be effectively realized by terminating the sample collection process upon reaching the desired number of false positives. We choose the number of false positive finds to be either 0 or 10 or the average maximal number of false positives found by other methods for one dataset. These three cases are reasonable because 0 false positives are the desired outcome for our use case. 10 is the maximum allowed number of false positives due to the use case. The last is interesting for observations based on the other quality measurements used to compare all methods (see \Cref{sec:041_qualitMeasures}). To avoid the strong influence of an outlier that could happen by executing the sampling process only once per data set and several false positives, each process is repeated 100 times. The resulting quality measurements are then averaged before being used for comparison. This baseline can be seen as a simple method to show that a more complex method is necessary for this problem.

No further preprocessing steps were needed for the first four methods and the baseline before the experiment could be run. 

More steps needed to be taken before the experiment for the ML-assisted predict and compare method. As the ML method, \lstm{} is used. This \lstm{} needs to be trained before it can be used for prediction (details in \Cref{03_lstmCusum}).

For each method, multiple experiments were run with different parameter sets. How we defined the parameter ranges is explained in \Cref{subsec:03_parameter}.  The packages and parameters used in the experiments are described in the following subsections.

\subsubsection{\bfast}
For the implementation of the \bfast{} method, the eponymous package \bfast{} \cite{verbesselt_near_2012} was utilized. This package implements different functions. The \textit{bfastmonitor} function is the implementation of the online version of \bfast{} described in \cite{verbesselt_near_2012}.

For the following parameters $minHist$, $histFact$, $h$ and $level$ different values were tested in the experiments. The first two parameters define the length of the stable history accessible to \textit{bfastmonitor}. 
$histFact$ defines a percent value of the available history that is stable and $minHist$ ensures that there is at least a certain amount of history available to the function. The maximum amount of history was also limited but never changed during the experiments.

$h$ specifies the bandwidth of the mosum process. The range of $h$ is between 0 and 1, as the bandwidth should be defined relative to the data available. $level$ sets the probability that a type 1 error occurs. $level$ and $h$ are parameters used by \textit{bfastmonitor}.

\subsubsection{Bayesian }
The experiments on the Bayesian online method were conducted with the online CPD method found in the ocp \cite{AdamsMacKay} package. During the experiments the threshold parameter called $cpthreshold$ was varied. This parameter defines the value of the run length probability, which, if exceeded, leads to a change point being detected. 

\subsubsection{\ocd}
For \ocd{}, an implementation in an R Package of the same name exists. There, the sensitivity of the detector is regulated by two thresholds {\em diag} and {\em offDiag}, one for the diagonal statistic and one for the off-diagonal statistic calculated by \ocd, respectively. For the change point detection, the $getData$ function is used. This function calculates the statistics and checks if the results are above (change point found) or below the chosen threshold. The package further provides a method to estimate the parameters necessary for the calculations. This is done at the beginning and after each found change point.

\subsubsection{\cusum}
For our experiments with the \cusum{} change point detection method, the \cusum{} function from the qcc package \cite{qcc_2004} was used. For this change point detection method, only one parameter was tested with different values, the $decision.interval$ ($desInt$). This parameter is the same as the $\lambda$ parameter named in \Cref{subsec:02_class_CUSUM}, the qcc package just uses a different name($decision.interval$). The $decision.interval$ controls the sensitivity of \cusum. A high value for the 
{\em decision.interval} leads to few change points detected and a low value to many detected change points. 

\subsubsection{\arima{} \cusum}\label{sec:arimaCusum}
Here, \cusum{} from qcc was also used. For the \arima{} part, the implementation from the forecast \cite{forecast_2008,forecast_2022} package was used. The function used is called \textit{auto.arima}, this method decided automatically the order of the \arima{} model that fits the data provided best. As we work with heterogeneous data, where different wear behaviors can be observed in different experiment stages, it was decided to let the auto.arima function fit the suitable model for the data. 
This fit is only performed at the beginning and after each detected change point, as at those points, we need to fit the model to a new behavior. Furthermore, fitting the \arima{} model only at those points leads to faster results.
The \arima{} part adds no parameters that need to be considered during the experiments, therefore only the $decision.interval$ ($desInt$) value is varied during the experiments.

\subsubsection{\lstm{} \cusum}
\label{03_lstmCusum}
The qcc implementation of the \cusum{} method was used once more. The implementation of the \lstm{} part was done by utilizing the well-known Keras \cite{kerasR_2017} package for R. \textit{\lstm\_model\_fit}, \textit{predict} and \textit{modelFit} are only some of the functions used for the implementation, additionally to the $decision.interval$ ($desInt$), two more parameters must be considered during the experiment. The length of the input window (number of input neurons for the \lstm)($nh$) and the size of the future window (number of output neurons for the \lstm)($nz$). 
Differently from all the other methods mentioned above, the \lstm{} needed to be trained before it could be used successfully for a prediction and in an experiment. Therefore, it was necessary to prepare training data from tribological experiments without divergent wear. 
This choice also helps to prevent overfitting, as the data sets used for training are different from those in the experiments. Due to the missing divergent wear part, these data sets are not used to evaluate the change point detection methods.
 The \lstm s were trained with the z score standardized training data sets. The best number of epochs and the batch size were determined by comparing the results of the prediction accuracy for a training and validation data set. The predictive power of the \lstm{} has not been optimized over the number of training samples.
 However, the size was chosen to be large enough (500) to recognize the typical trend patterns before the specific CPs (run-in behavior, linear increase of steady-state regime). The trained \lstm s were then used together with \cusum{} in the experiments to detect change points in the tribological data with divergent wear (Table 3), and constant wear after the run-in (Table 6).

\subsection{Parameter selection for Experiment}
\label{subsec:03_parameter}

For the experiments, different ranges of parameters for each method were used. For all the parameter ranges, the same criteria apply. Inside the chosen range, all settings have to find at least one change point for at least one dataset and less than 1000 change points for at least one data set. We tested the different methods with extreme parameter values to find those borders. After defining the borders, step sizes for each range were defined. Every step corresponds with a parameter value that was tested. For those methods with multiple parameters (\bfast{} and \lstm{} \cusum), the experiment was conducted for each parameter value combination. 

Furthermore, some of these parameters (nh, nz and minHist) are responsible for the amount of data the method has access to. For these parameters, another criterion for the boundary was considered. Due to the nature of the data, the first 600 data points do not have any change points. Therefore, they can be used safely as historical data without the risk of overlooking a change point.

\section{Results and Comparison}
\label{sec:04_res_comp}

This section focuses on the results of the experiments described in \Cref{sec:exp_disc}. In \Cref{04_Result_exp} the results of the different experiments are depicted. A discussion about the results of \pandc{} compared to the results of the reference methods can be found in \Cref{subsec:04_disc}. This is followed by a discussion of how different parameter settings influence the results in \Cref{subsec:04_param}. In the Beginning (\Cref{sec:041_qualitMeasures}), the Metrics used to gauge the quality of the results are described.

\subsection{Quality measures}
\label{sec:041_qualitMeasures}

This chapter shows the results of the experiments described in \Cref{sec:exp_disc}. We first note that the type I error measure is typically $\textrm{ARL}_0$, the average run length between two false positive detections (average run length in control). The type II error is usually measured by the average time between the actual occurrence and detection of a change point (average run length out of control, denoted by $\textrm{ARL}_\Delta$, \cite{hinkley70}). Note that for unlabeled data, the estimator of this measure is usually defined by the time between the estimated occurrence of the change point and its detection time, which entails another source of uncertainty versus our case of known ground truth change point positions in time.

Instead of using $\textrm{ARL}_0$, we use the number of false positive events (detections which do not correspond to real change points (labeled by experts according to the definition given in \Cref{subsec:03_dataExp})) before the occurrence of a specific change point, which will be called \fpc. This choice arises from the notable variations in observation counts across our datasets (see \Cref{tab:04_dataLen}), rendering the actual count of false positives a more sensible gauge of quality than an associated probability-expressing ratio. 

\begin{table}[ht]
\centering
\label{tab:04_dataLen}
\caption{This shows an overview of the length of the different wear phases in each dataset and the total length of each dataset. It can be seen that the length of the phases varies significantly between the different datasets.}
\begin{tabular}{rrrrrr}
  \hline
 dataset & \begin{tabular}[c]{rr}running-in \\ wear \end{tabular}& \begin{tabular}[c]{rr}constant \\ wear \end{tabular} & \begin{tabular}[c]{rr}divergent \\ wear \end{tabular} & \begin{tabular}[c]{rr}experiment \\ paused \end{tabular} & all Datapoints \\ 
  \hline
 1 & 4005 & 13596 &  25 & 2010 & 19636 \\ 
 2 & 0 & 4239 & 6605 & 96871 & 107715 \\ 
 3 & 802 & 1842 & 3158 & 10019 & 15821 \\ 
 5 & 0 & 29237 & 15515 & 327592 & 372344 \\ 
 7 & 0 & 3191 & 5233 & 20351 & 28775 \\ 
   \hline
\end{tabular}
\end{table}

For the type II error, we employ an estimator based on $\textrm{ARL}_\Delta$. The number of discrete time steps between the label and the detection is transformed into a percentage value to increase the comparability between the different datasets. The basic value for the calculation is the number of data points of the wear phase, which begins with the labeled change point corresponding to the found change point. 
\Cref{fig:04_metricPlot} shows the result of a change point detection method applied to dataset number 3. To calculate the \arlp{} of the found change point between the constant and the divergent wear phase (found CP K$-$\textgreater A) for this example, one needs the position of the change point (3476) and the length of the divergent wear phase (3158), which is defined by two labeled change points, one between the constant and the divergent wear phase (labeled CP K$-$\textgreater A) and one between the divergent wear and a pause in the experiment (labeled CP A$-$\textgreater V). Therefore, the \arlp{} is 7.32. 
The \fpc{} for this change point is two because the two additional false positive detections after the detected change point are not counted due to the online setting of the scenario in which the experiment is stopped in case of the detection of this change point. Therefore, those change points would not be detected. 

The following applies to all discussed results. The \arlp{} and \fpc{} were only calculated for the results of runs that found at least the change in the divergent wear. 

\begin{figure}[htbp]
    \centering
 \includegraphics[width=1\linewidth, height=5cm]{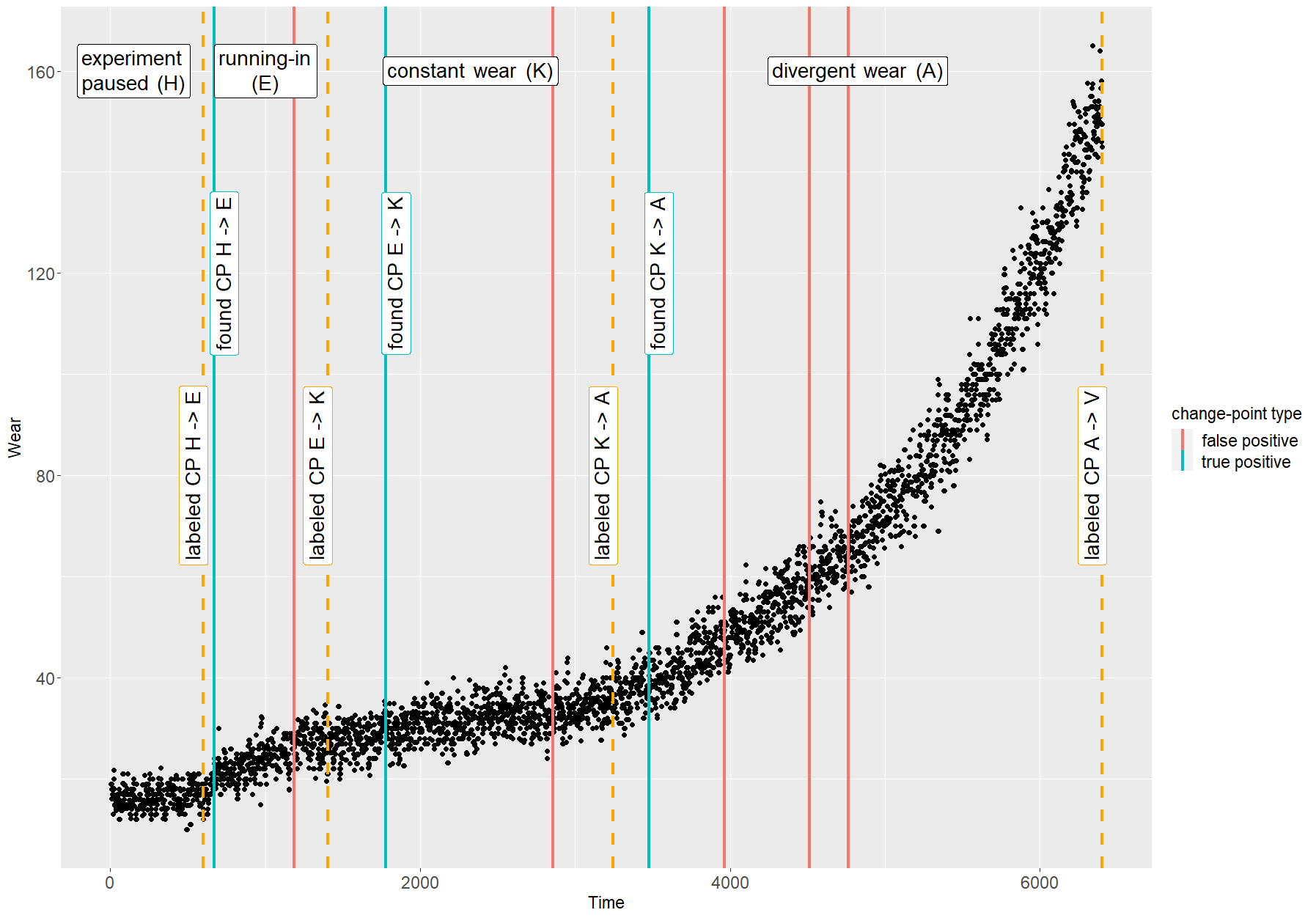}
    \caption{ The multiple vertical lines indicate either change points labeled by experts (black), true positive (blue) or false positive (red) detections, in dataset number 3.}
    \label{fig:04_metricPlot}
\end{figure}

\subsection{Results}
\label{04_Result_exp}

\Cref{tab:04_minMaxResAll} gives an overview in which each method is independently summarized in terms of its best and worst results. A first look at these values shows for the false positives that

\begin{itemize}
    \item for at least one parameter set, \arima{} \cusum, \cusum{}, \lstm{} \cusum{} and \ocd{} produced very few or non false-positives;
    \item looking at the worst-case scenarios regarding the Fp criteria, \arima{} \cusum{} is still the method with the least amount of false positives;
    \item another interesting observation is that all method have their highest Fpc for the $5^{th}$ dataset. 
\end{itemize} 
\hspace{-0.1cm} In terms of the ArlP, 

\begin{itemize}
    \item  \ocd{} is the fastest method to detect change points, closely followed by \bfast{}, \lstm{} \cusum, \cusum{} and Bayesian.
    \item The fastest result for the \pandc methods (\arima{} \cusum{} and \lstm{} \cusum) are reached for data set 3. In contrast, the other methods all have their fastest results at data set 7 (except for \ocd{} for which data set 5 delivers the fastest result).
\end{itemize}

Looking at the minimal and maximal values of the criteria separately gives an understanding of the boundaries of the method’s performance. The separate consideration of the criteria means that the lowest value of one criterion does not automatically correspond with the lowest criteria of the other. To find the best-performing method, it is essential to find the run with the best combination of Fpc and ArlP. 
The best combination of these two metrics depends on the specific scenario in which the change point detection is used. In some scenarios, a low Fpc is more important than a low ArlP. In others, the fast detection of change points is so important that a higher Fpc is acceptable. In the scenario we look at in this paper, a low number of false positives is important (up to 10 is acceptable), even if it might lead to a longer detection time.

\definecolor{LightGray}{rgb}{0.83,0.83,0.83}
\definecolor{LightLightGray}{rgb}{0.95,0.95,0.95}

\begin{table}[ht]
\centering
\caption{This table shows the minimal and maximal value of the false-positive count (Fpc) and the average run length percent (ArlP) per method and dataset.} 
\label{tab:04_minMaxResAll}
\begin{tabular}{lrrrrr}
  \rowcolor{LightGray} \hline
Method & Dataset & min Fp & max Fp & min ArlP & max ArlP \\ 
  \hhline{======} \hline
   &   1 &   0 &  20 & 30.50 & 82.45 \\ 
   &   2 &   0 &  17 & 18.76 & 70.75 \\ 
   &   3 &   0 &  10 & 1.11 & 63.18 \\ 
   &   5 &   0 & 375 & 14.70 & 78.02 \\ 
  \multirow{-5}{*}{\arima{} \cusum} &   7 &   0 &  17 & 21.48 & 88.80 \\ 
   \rowcolor{LightGray} &   1 &  0 & 169 & 44.5 & 48.4 \\ 
   \rowcolor{LightGray} &   2 &  0 & 37 & 9.12 & 46.9 \\ 
   \rowcolor{LightGray} &   3 &  0 & 124 & 3.16 & 49.9\\ 
   \rowcolor{LightGray} &   5 &  0 & 2136 & 0.71 & 25.1 \\ 
   \rowcolor{LightGray}\multirow{-5}{*}{Baseline} &   7 &   0 & 152 & 3.68 & 42.6 \\ 
   &   1 &  &  &  &  \\ 
   &   2 & 218 & 1091 & 1.95 & 7.66 \\ 
   &   3 & 102 & 255 & 3.99 & 10.10 \\ 
   &   5 & 581 & 2135 & 0.10 & 0.10 \\ 
  \multirow{-5}{*}{Bayesian} &   7 &  81 & 338 & 0.06 & 0.63 \\ 
   \rowcolor{LightGray} &   1 &   0 & 118 & 2.52 & 98.44 \\ 
   \rowcolor{LightGray} &   2 &   5 & 265 & 0.72 & 64.17 \\ 
   \rowcolor{LightGray} &   3 &   0 &  18 & 0.16 & 80.66 \\ 
   \rowcolor{LightGray} &   5 &  39 & 2282 & 0.01 & 55.68 \\ 
   \rowcolor{LightGray}\multirow{-5}{*}{\bfast} &   7 &  15 &  97 & 0.00 & 75.33 \\ 
   &   1 &   2 & 1044 & 2.52 & 70.46 \\ 
   &   2 &   0 & 884 & 0.72 & 38.97 \\ 
   &   3 &   0 & 185 & 0.06 & 21.38 \\ 
   &   5 &   0 & 8101 & 0.13 & 77.65 \\ 
  \multirow{-5}{*}{\cusum} &   7 &   0 & 830 & 0.00 & 78.50 \\ 
   \rowcolor{LightGray} &   1 &   4 & 236 & 2.52 & 78.46 \\ 
   \rowcolor{LightGray} &   2 &   0 & 619 & 0.41 & 92.82 \\ 
   \rowcolor{LightGray} &   3 &   0 & 123 & 0.19 & 92.03 \\ 
   \rowcolor{LightGray} &   5 &   0 & 1235 & 1.94 & 77.48 \\ 
   \rowcolor{LightGray} \multirow{-5}{*}{\lstm{} \cusum} &   7 &   0 & 151 & 0.61 & 91.63 \\ 
   & 1 & 0 & 158 & 0.06 & 90.4 \\
   & 2 & 0 & 79 & 0.02 & 95.2 \\
   & 3 & 0 & 29 & 0.10 & 97.1\\
   & 5 & 0 & 3379 & 0.01 & 94.8\\
  \multirow{-5}{*}{\ocd}  & 7 & 0 & 145 & 0.44 & 79.6\\
   \hline
\end{tabular}
\end{table}

\Cref{tab:04_bestOne10} shows the best runs per dataset and method where the Fpc was less than or equal to 10. In case more than one candidate per data set and method was found, the following rule was applied: First, the smallest Fpc and then the smallest ArlP. ((\Cref{tab:04_bestOne10}) show the result for the opposite rule). The result shows that

\begin{itemize}
    \item for datasets 2, 3, 5 and 7 the \lstm{} \cusum{} is the best performing one;
    \item for dataset 1, \bfast{} is the best performing one;
    \item \arima{} \cusum, \cusum{},\ocd{} and \lstm{} \cusum{} have at least one parameter setting that works well for each dataset. 
\end{itemize}

The difference between dataset 1 and the other datasets lies in the very short divergent wear phase (25 data points) compared to the others. 

\begin{table}[ht]
\centering
\caption{This Table shows the best result per dataset and method under the condition that 10 false-positives were found at max. Change point considered here (only): Divergent Wear.} 
\label{tab:04_bestOne10}
\begin{tabular}{lrrrl}
  \rowcolor{LightGray} \hline
Method & Dataset & Fpc & ArlP & Parameter \\ 
  \hhline{=====} \hline
  \bfast & \multirow{5}{*}{1} & 0 & 14.51 & \makecell[l]{minHist: 175, histFact: 0.30, \\ h: 0.50, level: 0.001} \\ 
  \arima{} \cusum &    & 0 & 38.49 & desInt: 100 \\ 
  \ocd & & 0 & 42.5 & diag: 210001, offDiag: 4100001 \\
  \cusum &    & 2 & 26.50 & desInt: 400 \\ 
  \lstm{} \cusum &    & 4 & 78.46 & desInt: 400, nh: 100, nz: 20 \\ 
   \rowcolor{LightGray} \lstm{} \cusum &  & 0  & 1.94 & desInt: 300, nh: 500, nz: 100 \\ 
   \rowcolor{LightGray}\cusum &    & 0 & 4.72 & desInt: 400 \\ 
   \rowcolor{LightGray}\arima{} \cusum &    & 0 & 18.69 & desInt: 350 \\ 
   \rowcolor{LightGray}\ocd & & 0 & 36.1 & diag: 4201, offDiag: 156001 \\
   \rowcolor{LightGray}Baseline & & 0 & 52.0 & \\
   \rowcolor{LightGray}\bfast &  \multirow{-6}{*}{2}  & 5 & 64.04 & \makecell[l]{minHist: 175, histFact: 0.25, \\ h: 0.50, level: 0.005} \\ 
  \lstm{} \cusum &   \multirow{5}{*}{3} & 0 & 0.22 & desInt: 120, nh: 600, nz: 50 \\ 
  \ocd & & 0 & 1.71 & diag: 101, offDiag: 501 \\
  \cusum &    & 0 & 5.29 & desInt: 130 \\ 
  \arima{} \cusum &    & 0 & 22.99 & desInt: 200 \\ 
  Baseline & & 0.1 & 43.4 & \\
  \bfast &    & 0 & 80.66 & \makecell[l]{minHist: 50, histFact: 0.25,\\ h: 0.50, level: 0.002} \\ 
   \rowcolor{LightGray} \lstm{} \cusum &  & 0 & 48.39 & desInt: 125, nh: 600, nz: 40 \\ 
   \rowcolor{LightGray}Baseline & & 2.35 & 50.0 & \\
   \rowcolor{LightGray}\arima{} \cusum &    & 0 & 51.78 & desInt: 450 \\ 
   \rowcolor{LightGray} \ocd & & 0 & 53.1 & diag: 210001, offDiag: 4100001 \\
   \rowcolor{LightGray}\cusum & \multirow{-5}{*}{5} & 0 & 77.65 & desInt: 300 \\ 
  \lstm{} \cusum & \multirow{3}{*}{7} & 0 & 14.56 & desInt: 140, nh: 600, nz: 200 \\ 
  \ocd & & 0 & 15.3 & diag: 4201, offDiag: 6001 \\
  \arima{} \cusum &    & 0 & 36.54 & desInt: 325 \\ 
  \cusum &    & 0 & 37.07 & desInt: 150 \\ 
  Baseline & & 1.11 & 43.9 & \\
   \hline
\end{tabular}
\end{table}

The next question is to find the one parameter set that works best for all data sets combined. According to \Cref{tab:04_bestAll150}, in this sense
\begin{itemize}
    \item \arima{} \cusum{} has the {\bf  best overall} performance;
    \item the best overall result for \lstm{} \cusum{} has a larger Fpc than $50$. 
\end{itemize}

The Fpc and ArlP values for each parameter set and method were summarised to generate this table. Therefore,
The threshold of 50 comes from the Fpc threshold (10) initially being multiplied by the number of data sets (5). However, in order to also include a \lstm{} \cusum{} result, the threshold used in \Cref{tab:04_bestAll150} was shifted further up to 150. \\


\begin{table}[ht]
\centering
\caption{Results grouped by parameter set over all datasets. 
With 150 false positives found at max, at least the divergent-wear change was found.} 
\label{tab:04_bestAll150}
\begin{tabular}{lrrrrl}
  \rowcolor{LightGray} \hline
Method & $\Sigma$ Fpc & sd Fpc & $\Sigma$ ArlP & Sd ArlP & Parameter \\ 
  \hhline{======} \hline
\arima{} \cusum &   1 & 0.45 & 252.94 & 17.98 & desInt: 725 \\ 
   \rowcolor{LightGray}\arima{} \cusum &   1 & 0.45 & 287.47 & 20.99 & desInt: 700 \\ 
  \arima{} \cusum &   2 & 0.89 & 265.22 & 16.73 & desInt: 650 \\ 
   \rowcolor{LightGray}\arima{} \cusum &   2 & 0.89 & 269.23 & 17.38 & desInt: 675 \\ 
  \arima{} \cusum &   4 & 1.10 & 187.53 & 17.09 & desInt: 575 \\ 
   \rowcolor{LightGray}\arima{} \cusum &   4 & 1.10 & 187.57 & 17.09 & desInt: 600 \\ 
  \arima{} \cusum &   4 & 1.10 & 225.57 & 15.40 & desInt: 625 \\ 
   \rowcolor{LightGray}\cusum &   4 & 1.79 & 237.08 & 28.96 & desInt: 300 \\ 
  \arima{} \cusum &   7 & 1.14 & 218.48 & 13.64 & desInt: 525 \\ 
   \rowcolor{LightGray}\arima{} \cusum &   7 & 1.14 & 218.48 & 13.64 & desInt: 550 \\ 
  \cusum &   8 & 2.61 & 179.75 & 27.35 & desInt: 200 \\ 
   \rowcolor{LightGray}\arima{} \cusum &   8 & 1.52 & 225.56 & 19.27 & desInt: 500 \\ 
  \cusum &  10 & 2.92 & 142.47 & 18.94 & desInt: 180 \\ 
   \rowcolor{LightGray}\cusum &  16 & 3.56 & 93.56 & 13.34 & desInt: 160 \\ 
   \cusum &  29 & 7.50 & 91.55 & 13.18 & desInt: 140 \\ 
  \rowcolor{LightGray}Baseline & 29.3 & 4.99 & 235 & 18.6 & \\
  \hline
   \cusum &  62 & 18.46 & 70.65 & 12.92 & desInt: 120 \\ 
  \rowcolor{LightGray}\lstm{} \cusum & 143 & 34.52 & 146.26 & 23.08 & \makecell[l]{desInt: 120, \\nh: 400, nz: 50} \\ 
   \rowcolor{LightGray} \hline
\end{tabular}
\end{table}



\begin{table}[ht]
\centering
\caption{Results grouped by parameter set over the datasets 3,5 and 7 with 30 false positives found at max.
Change Point: Divergent-wear.} 
\label{tab:04_bestOf3Datasets}
\begin{tabular}{lrrrrl}
  \rowcolor{LightGray} \hline
Method & $\Sigma$ Fpc & sd Fpc & $\Sigma$ ArlP & Sd ArlP & Parameter \\ 
  \hhline{======} \hline
\cusum & 0 & 0.00 & 175.60 & 33.85 & desInt: 300 \\ 
   \rowcolor{LightGray}\arima{} \cusum & 1 & 0.58 & 118.17 & 11.22 & desInt: 725 \\ 
  \arima{} \cusum & 1 & 0.58 & 156.82 & 27.42 & desInt: 700 \\ 
   \rowcolor{LightGray}\cusum & 2 & 1.15 & 99.09 & 24.27 & desInt: 180 \\ 
  \cusum & 2 & 1.15 & 135.29 & 33.51 & desInt: 200 \\ 
   \rowcolor{LightGray}\arima{} \cusum & 2 & 1.15 & 142.71 & 21.13 & desInt: 650 \\ 
  \arima{} \cusum & 2 & 1.15 & 142.73 & 21.13 & desInt: 675 \\ 
  \rowcolor{LightGray}Baseline & 3.57 & 1.13 & 137 & 3.69 & \\
   \arima{} \cusum & 4 & 1.15 & 109.08 & 13.00 & desInt: 625 \\ 
  \rowcolor{LightGray}\arima{} \cusum & 4 & 1.15 & 110.11 & 13.82 & desInt: 575 \\ 
  \arima{} \cusum & 4 & 1.15 & 110.14 & 13.82 & desInt: 600 \\ 
  \rowcolor{LightGray}\arima{} \cusum & 6 & 1.00 & 110.14 & 13.76 & desInt: 525 \\ 
   \arima{} \cusum & 6 & 1.00 & 110.14 & 13.76 & desInt: 550 \\ 
  \rowcolor{LightGray}\cusum & 7 & 4.04 & 55.17 & 16.73 & desInt: 160 \\ 
   \arima{} \cusum & 7 & 1.53 & 104.32 & 15.46 & desInt: 500 \\ 
  \rowcolor{LightGray}\lstm{} \cusum & 11 & 4.04 & 133.92 & 12.58 & \makecell[l]{desInt: 120, \\nh: 400, nz: 50} \\ 
   \lstm{} \cusum & 15 & 5.00 & 80.73 & 22.79 & \makecell[l]{desInt: 125, \\nh: 500, nz: 50} \\ 
   \rowcolor{LightGray} Baseline & 17.8 & 4.74 & 102 & 6.10 & \\
  \cusum & 19 & 10.12 & 53.77 & 16.64 & desInt: 140 \\ 
   \rowcolor{LightGray} \lstm{} \cusum & 19 & 7.09 & 80.73 & 22.79 & \makecell[l]{desInt: 120, \\nh: 500, nz: 50} \\ 
  \lstm{} \cusum & 24 & 8.00 & 150.74 & 27.23 & \makecell[l]{desInt: 100, \\nh: 500, nz: 40} \\ 
   \rowcolor{LightGray} \lstm{} \cusum & 25 & 10.41 & 86.45 & 21.54 & \makecell[l]{desInt: 110, \\nh: 400, nz: 50} \\ 
   \ocd{} & 26 & 15.01 & 88.54 & 18.11 & \makecell[l]{diag: 4201, \\offDiag: 100000000}\\
   \rowcolor{LightGray} \ocd{} & 29 & 8.50 & 136.87 & 29.75 & \makecell[l]{diag: 1201,\\ offDiag: 100000000}\\
   \hline
\end{tabular}
\end{table}



By restricting our observations to data sets 3, 5, and 7 we obtain the results of \Cref{tab:04_bestOf3Datasets} in which correspondingly an Fpc of at most 30 was allowed. We observe that

\begin{itemize}
    \item there are more results (due to fewer Fpc exceeding the threshold of 30);
    \item \cusum{} wins, but \arima{} \cusum{} is comparable in its performance.
    \item \lstm{} \cusum{} is included within the cases with Fpc counts less than 30.
\end{itemize}

This allows recognizing data sets 3, 5, and 7 similar in the sense of methods with single parameters sets being universally applicable on them.\\

Finally, we observe for the change point into the steady-state wear regime (instead of the divergent wear regime) we see from \Cref{tab:04_bestResForK} that 

\begin{itemize}
    \item the Predict and Compare Methods are the most successful on data sets 1 and 5;
    \item \arima{} \cusum{} always appears in the cases with a Fpc of less than 10.  
\end{itemize}

The fact that in this table (\Cref{tab:04_bestResForK}) some methods do not appear in the lost of specific data sets is related to the fact that the change point into the steady-state wear regime does not occur for some parameter sets from \Cref{tab:04_bestOne10}.

\begin{table}[ht]
\centering
\caption{This Table shows the best result per dataset and method under the condition that 10 false-positives 
where found at max. Change point considered here (only): Constant Wear.} 
\label{tab:04_bestResForK}
\begin{tabular}{lrrrl}
  \rowcolor{LightGray} \hline
Method & Dataset & Fpc & ArlP & Parameter \\ 
  \hhline{=====} \hline
\arima{} \cusum &    &   0 & 3.61 & desInt: 75 \\ 
\ocd{} & & 0 & 20.0 & diag: 4201, offDiag: 156001 \\
  Baseline &     & 0.17 & 42.7  & \\ 
  \bfast &    &   1 & 0.40 & \makecell[l]{minHist: 175, histFact: 0.30, \\h: 0.50, level: 0.002} \\ 
  \cusum &    &   1 & 14.14 & desInt: 300 \\ 
  \lstm{} \cusum &    & 7 & 67.40 & desInt: 300, nh: 50, nz: 100 \\ 
  Baysian &   \multirow{-7}{*}{1} &   9 & 3.74 &  \\ 
   \rowcolor{LightGray}\ocd{} & & 0 & 8.87 & diag: 501, offDiag: 501 \\
   \rowcolor{LightGray}\cusum &    &   0 & 9.55 & desInt: 110 \\ 
   \rowcolor{LightGray}\arima{} \cusum &    &   0 & 18.45 & desInt: 125 \\ 
   \rowcolor{LightGray}Baseline &    & 0.1 & 50.2 & \\
   \rowcolor{LightGray}\bfast &   \multirow{-6}{*}{2} &   3 & 0.19 & \makecell[l]{minHist: 200, histFact: 0.30, \\h: 0.50, level: 0.005} \\ 
   \ocd{} & & 0 & 0.69 & diag: 1201, offDiag: 4201 \\
  \arima{} \cusum &    &   0 & 20.77 & desInt: 325 \\ 
  Baseline &    & 0.03 & 47.4 & \\ 
  \bfast &    &   2 & 2.05 & \makecell[l]{minHist: 225, histFact: 0.30, \\h: 0.50, level: 0.001} \\ 
  \cusum &    &   3 & 37.27 & desInt: 80 \\ 
  \lstm{} \cusum &   \multirow{-6}{*}{3} &   6 & 13.50 & desInt: 100, nh: 200, nz: 50 \\ 
   \rowcolor{LightGray} \lstm{} \cusum &    &   0 & 3.23 & desInt: 110, nh: 500, nz: 40 \\ 
   \rowcolor{LightGray}\cusum &    &   0 & 5.58 & desInt: 300 \\ 
   \rowcolor{LightGray}Baseline &    & 2.26 & 42.4 & \\ 
   \rowcolor{LightGray}\arima{} \cusum &   \multirow{-3}{*}{5} &   4 & 94.04 & desInt: 150 \\ 
  \ocd{} & & 0 & 62.5 &  diag: 4201, offDiag: 4201 \\
  \arima{} \cusum &  &   0  & 71.64 & desInt: 250 \\ 
  \cusum &  &   0 & 74.65 & desInt: 150 \\ 
  Baseline &    & 1.03 & 56.1 & \\ 
  \bfast &  &   2 & 67.97 & \makecell[l]{minHist: 200, histFact: 0.25, \\h: 0.50, level: 0.001} \\ 
  \lstm{} \cusum & \multirow{-6}{*}{7} &   2 & 88.97 & desInt: 175, nh: 200, nz: 100 \\ 
   \hline
\end{tabular}
\end{table}



\vspace{0.3cm}

\subsection{Discussion}
\label{subsec:04_disc}

We first note that \pandc\, as a fully online working algorithm, it can -in principle- handle data needing large prediction windows (cf. large $\varepsilon$-values in the $\varepsilon$-realtime algorithms in \cite{amin_2017}) such as those with gradual changes (slow onsets of trend-changes) well, in the sense that the time until detection (and therefore, on average, the Arl) even for CPs occurring at the beginning of these windows can be small and not on the order of magnitude of the prediction window size. Furthermore, since hopping windows are used (not sliding), the multiple testing problem of sliding windows with non-empty intersection is avoided and taken care of in the sense of the sequential CUSUM test. In this sense, there are no problems of ambiguous results of different CP-results of overlapping sliding windows.

Therefore, we arrive at noting that with the right tuning, Predict and Compare works well for change points with {\bf gradually developing onsets} of the anomalies (cf. with the Z-score of data set 3, \Cref{fig:allData_Z}). In particular, for specific data sets and appropriately adjusted parameters \lstm{} \cusum{} is the best method when it is important to keep the number of false positive detections at a minimum (below 10). Furthermore, \arima{} \cusum{} succeeds in the additional constraint of using the same parameter set when applying the same method to different data sets. 

More precisely, the emphasized feature B. in Section 1 requires few false positives and finding the change point quickly even though the onset develops gradually. The result that LSTM CUSUM is best for individually tuned parameters (Table 3) and ARIMA CUSUM is the best method if the same parameters are used throughout the different data sets (Table 2) shows the heavy sensitivity of the method with respect to correct parameter-tuning. This lack of parameter-robustness when using an advanced perdictive model must be seen as a weakness of the method. 

Utilizing a generated sample as shown in \Cref{fig:04_genData_cusum}, this can be visualized. Part B shows the result of the Predict and Compare method applied to the generated sample. In part C and D, the results of a standard \cusum{} can be seen. Part B and part C have the same threshold and detect the change point at the same time. However, part B has no false positives, in contrast to part C, which has several. Part D has no false positives like part B but detects the change point later than the method used in part B.

\begin{figure}[htbp]
    \centering
 \includegraphics[width=1\linewidth, height=6.5cm]{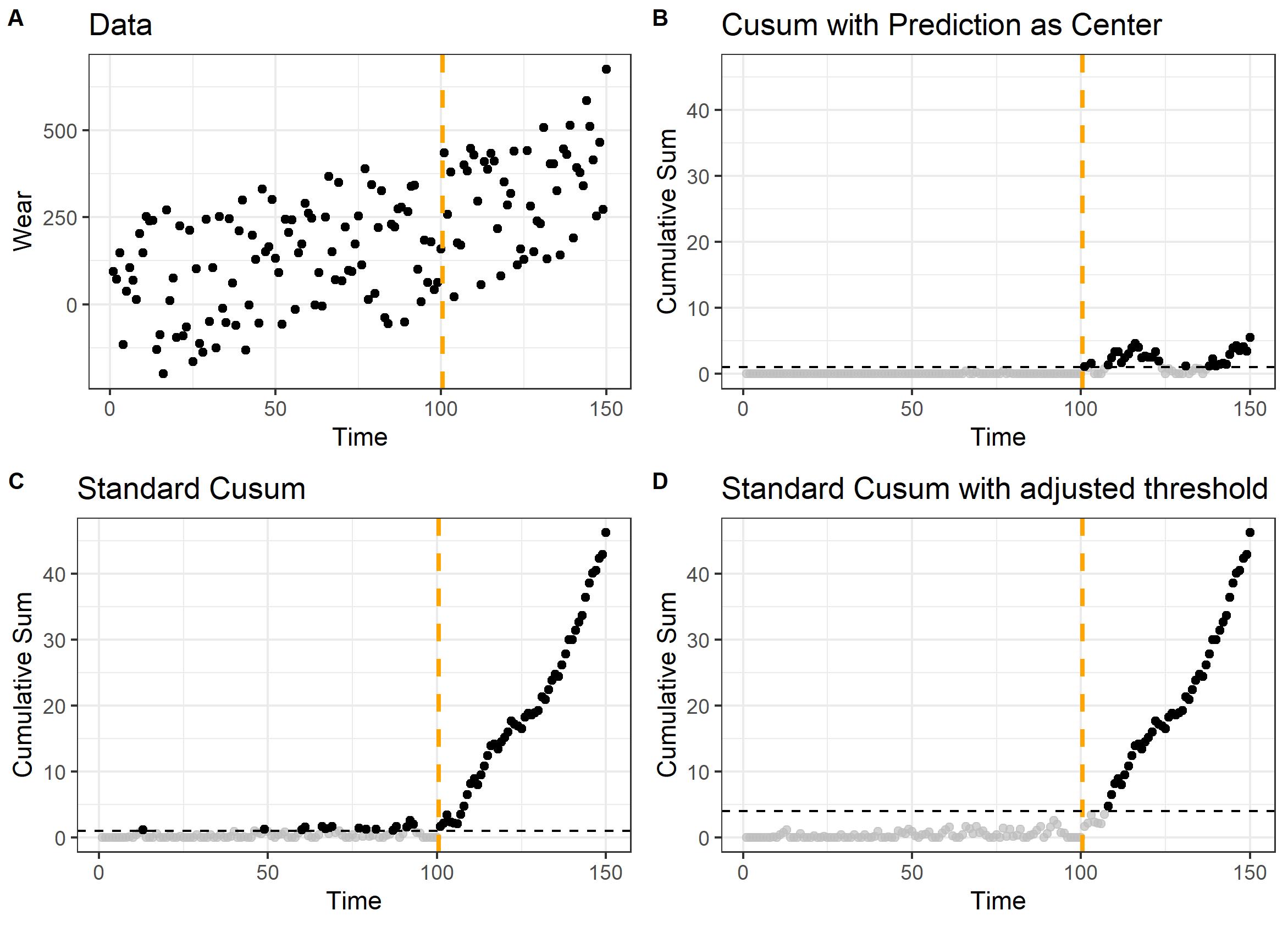}
    \caption{Part A shows some generated data with a change point at 101. Part B,C and D show results using the \cusum{} method with different thresholds and $\theta$. Part B using a prediction as $\theta$ and parts C and D the standard $\theta$ of \cusum. The thresholds of part B and C are the same and the threshold of D is higher. The black dots in part B-D represent those times when a change is detected by \cusum. The grey dots represent those times when no changes is picked up by \cusum. The horizontal dashed line is the threshold (in part B-D). The vertical black line in all the plots visualizes the real change point.  It is seen that in standard CUSUM the same lack of false positives can only be achieved with a higher (less sensitive) threshold. }
    \label{fig:04_genData_cusum}
\end{figure}

\subsection{Parameter}
\label{subsec:04_param}

In \Cref{subsec:03_impl} and \Cref{subsec:03_parameter}, we discuss the different parameters used for our tests. Here, we share our observations on how different parameter values influence the quality metrics ArlP and Fpc. 

\arima{} \cusum, \cusum{} and Bayesian share a commonality: each has one parameter to consider. 
The impact of varying $desInt$ values on ArlP and Fpc quality measures for \cusum{} is illustrated in \Cref{fig:04_param_ArlP_Fp_CUSUM}, revealing a divergent evolution of ArlP and Fpc. Specifically, as $desInt$ increases, Fpc exhibits a declining trend, whereas ArlP displays an ascending pattern. Since the parameter $desInt$ in \arima{} \cusum{} originates from the \cusum{} part, the observable behavior is the same.
While the parameter of Bayesian differs from that of \cusum{} ($desInt$), there is an observable trend where Fpc decreases with higher $cpthreshold$, while ArlP increases with the same increasing $cpthreshold$.
\ocd{} involves the consideration of two thresholds; however, due to the one-dimensional nature of our data, opting for a sufficiently high value for one threshold renders the influence on the outcome primarily on the other threshold. Consequently, the parallels with \cusum{} emerge, as the latter method operates with a solitary threshold.

\begin{figure}[ht]
    \centering
 \includegraphics[width=0.8\linewidth, height=6cm]{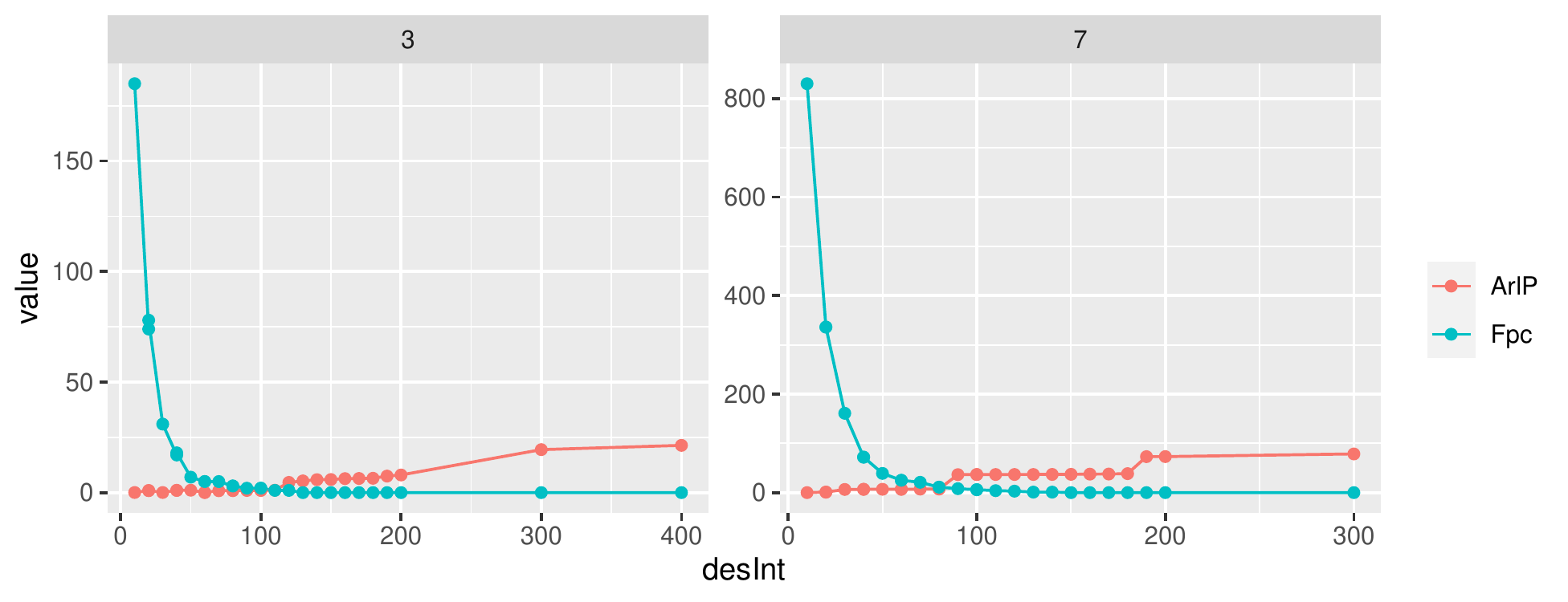}
    \caption{Each plot represents the development of ArlP and Fpc over different $desInt$ values for a certain dataset.}
    \label{fig:04_param_ArlP_Fp_CUSUM}
\end{figure}

For \lstm{} \cusum{} and \bfast, it is a bit more challenging to determine which parameter values will lead to which result, as there is more than one parameter to consider. \lstm{} \cusum{} has three parameters $nh$, $nz$ and $desInt$, which influence the Fpc an ArlP. If the $nz$ decreases, the Fpc decreases as well. An increase in the $nh$ parameter leads to a slight increase in the Fpc. To get a low Fpc value over multiple datasets, the middle range of the $nz$ and $nh$ parameters performed the best. The $desInt$ behaves similarly to the $desInt$ in \cusum{} and \arima{} \cusum{} and should be chosen as small as possible to reduce the amount of Fpc.

For our experiments, we used four different parameters for \bfast{} ($minHist$, $histFact$, $h$ and $level$). First, looking at each parameter separately, one can observe that the Fpc decreases with increasing $minHist$ and $h$. For $histFact$ and $level$, the opposite can be observed, with an increase in the parameter value, the Fpc also increases. For ArlP, the opposite is the case, low values for $minHist$ and $h$ lead to a low ArlP and high values to a high ArlP. Hence, low values for $histFact$ and $level$ lead to a high ArlP and high values to a low ArlP. These effects can still be observed by looking at the combinations of those different values.

In general, it shows that the well understood \cusum{} rule with its strong control of small type I error probabilities used in the comparison step of \pandc{} is critical for the goal of strong robustness against outliers followed here. Here, a different application, where detecting a specific type of change point is more important than a few false positives, may call for a different method in the comparison step. 

\section{Conclusion}
\label{05_Result_con}
To first  answer question {\bf Q3} and give  the conclusions directly referring to the experiment, we report that \pandc\, performs mostly better than the reference change point detection methods. Looking at the best results for each data set (\Cref{tab:04_bestOne10}) \lstm{} \cusum{} (data set 2,3,5 and 7) and \bfast{} (data set 1) are the best methods found by our experiments. The difference between datasets 2,3,5 and 7 and dataset 1 is the kind of divergent wear. A very short and steep slope characterizes the divergent wear part in data set 1. In contrast, the divergent wear in the other data set develops over a longer time period. The best method with a specific parameter set over all data sets is \arima{} \cusum{}. Therefore, the \arima{} \cusum{} approach is more generalized than the \lstm{} \cusum{} approach. But \arima{} \cusum{} takes longer until the change point is detected compared to \lstm{} \cusum{}. Hence, in the case of very similar data, \lstm{} \cusum{} is better. In the case of more diverse data, \arima{} \cusum{} is better. Therefore, we can - not surprisingly - conclude that a wider scope of applicability of the \pandc{} defined CPD detectors comes at the expense of the sophistication of the allowed intermediate trends. 

 Question {\bf Q2} about the natural use of predictive models to assist CPD is answered by Step (4.) of the Definition of \pandc\,(Section \ref{sec:pandc}), and exemplified by the observation of how non-trivial trends must be recognized and predicted before change points are distinctly recognized as such (and not mistaken with the trends). This can be seen in Table \ref{tab:04_bestOf3Datasets} (concerning the change point from the run-in into the steady-state wear regime): It is usually \arima{} \cusum{} or \lstm{} \cusum{} which are or rank among the best methods and often  outrank  \cusum{} in terms of the time until detection. The reason for this is well documented in \Cref{fig:04_genData_cusum}, which shows that \cusum{} only performs comparably to \cusum{} \lstm{} in terms of false positives, if the threshold is tuned upwards so much that a  significantly longer time until detection is observed.

 Finally, the answer to the more general {\bf Q1} is included in Definitions 1 and 2 (in \Cref{sec:defs_resq}) about trends and change points, and is illustrated in Figure \ref{fig:principle}:  The  trained predictive model recognizes the (curved) trend of the current (run-in) phase and predicts its continuation, facilitating the discovery of the discrepancy with the following trend (of the steady-state wear) regime.

We conclude that in accordance with our goals, Predict \& Compare provides a framework for the definition of ({\bf A}) online change point detectors, which ({\bf B}) allow detecting gradual structural changes and ({\bf C}) cope with trends that are not to be identified as change points. The two models used here in the prediction part of \pandc{} (\arima{}, \lstm{}) are  two well-known opposites in complexity and particularly useful in the tribological example, but by no means spanning the whole range of possible choices. Data $X_t$ of type (\ref{eq:Xrep}) referring to independent numbers of specific events occurring in time intervals with an intensity that changes slowly (as required by goal {\bf B}), the process has \pandc{} change points (Definition 2). Data with a richer auto-correlation structure may require a much longer predictive window size. Similarly, the \cusum{} rule is not the only thinkable comparison method between predicted and real values - especially if it is not the fluctuations around a trend that are the most relevant criterion for the comparison step. Thus, our demonstration of the effectiveness of \pandc{} in comparison with other state of the art online CPD methods only shows a few of a wide variety of different application scenarios.


\subsection*{Acknowledgements}
This work was funded by the Austrian COMET-Program (Project K2 InTribology1, no. 872176), and carried out at the Software Competence Center Hagenberg.

\subsection*{Conflicts of interest}
The authors have no conflicts of interest to declare that are relevant to the content of this article.

\subsection*{Data availability statement}
The data that support the findings of this study are available from AC$^2$T but restrictions apply to the availability of these data, which were used under licence for the current study, and so are not publicly available. Data are however available from the authors upon reasonable request and with permission of AC$^2$T.

\bibliography{CUSUMlib.bib}


\begin{thebibliography}{62}
\ifx \bisbn   \undefined \def \bisbn  #1{ISBN #1}\fi
\ifx \binits  \undefined \def \binits#1{#1}\fi
\ifx \bauthor  \undefined \def \bauthor#1{#1}\fi
\ifx \batitle  \undefined \def \batitle#1{#1}\fi
\ifx \bjtitle  \undefined \def \bjtitle#1{#1}\fi
\ifx \bvolume  \undefined \def \bvolume#1{\textbf{#1}}\fi
\ifx \byear  \undefined \def \byear#1{#1}\fi
\ifx \bissue  \undefined \def \bissue#1{#1}\fi
\ifx \bfpage  \undefined \def \bfpage#1{#1}\fi
\ifx \blpage  \undefined \def \blpage #1{#1}\fi
\ifx \burl  \undefined \def \burl#1{\textsf{#1}}\fi
\ifx \doiurl  \undefined \def \doiurl#1{\url{https://doi.org/#1}}\fi
\ifx \betal  \undefined \def \betal{\textit{et al.}}\fi
\ifx \binstitute  \undefined \def \binstitute#1{#1}\fi
\ifx \binstitutionaled  \undefined \def \binstitutionaled#1{#1}\fi
\ifx \bctitle  \undefined \def \bctitle#1{#1}\fi
\ifx \beditor  \undefined \def \beditor#1{#1}\fi
\ifx \bpublisher  \undefined \def \bpublisher#1{#1}\fi
\ifx \bbtitle  \undefined \def \bbtitle#1{#1}\fi
\ifx \bedition  \undefined \def \bedition#1{#1}\fi
\ifx \bseriesno  \undefined \def \bseriesno#1{#1}\fi
\ifx \blocation  \undefined \def \blocation#1{#1}\fi
\ifx \bsertitle  \undefined \def \bsertitle#1{#1}\fi
\ifx \bsnm \undefined \def \bsnm#1{#1}\fi
\ifx \bsuffix \undefined \def \bsuffix#1{#1}\fi
\ifx \bparticle \undefined \def \bparticle#1{#1}\fi
\ifx \barticle \undefined \def \barticle#1{#1}\fi
\bibcommenthead
\ifx \bconfdate \undefined \def \bconfdate #1{#1}\fi
\ifx \botherref \undefined \def \botherref #1{#1}\fi
\ifx \url \undefined \def \url#1{\textsf{#1}}\fi
\ifx \bchapter \undefined \def \bchapter#1{#1}\fi
\ifx \bbook \undefined \def \bbook#1{#1}\fi
\ifx \bcomment \undefined \def \bcomment#1{#1}\fi
\ifx \oauthor \undefined \def \oauthor#1{#1}\fi
\ifx \citeauthoryear \undefined \def \citeauthoryear#1{#1}\fi
\ifx \endbibitem  \undefined \def \endbibitem {}\fi
\ifx \bconflocation  \undefined \def \bconflocation#1{#1}\fi
\ifx \arxivurl  \undefined \def \arxivurl#1{\textsf{#1}}\fi
\csname PreBibitemsHook\endcsname

\bibitem{Lai_95}
\begin{barticle}
\bauthor{\bsnm{Lai}, \binits{T.L.}}:
\batitle{Sequential change point detection in quality control and dynamic systems}.
\bjtitle{J. Roy. Statist. Soc. (B)}
\bvolume{57},
\bfpage{613}--\blpage{658}
(\byear{1995})
\end{barticle}
\endbibitem

\bibitem{Wu_05}
\begin{bbook}
\bauthor{\bsnm{Wu}, \binits{Y.}}:
\bbtitle{Inference for Change Point and Post Change Means After a CUSUM Test (Lecture Notes in Statistics)},
\bedition{1}st edn.
\bpublisher{Springer},
\blocation{Heidelberg}
(\byear{2005}).
\burl{libgen.li/file.php?md5=72cfc291e96b7964b563672759bc9085}
\end{bbook}
\endbibitem

\bibitem{odile_2018}
\begin{bbook}
\bauthor{\bsnm{Pons}, \binits{O.}}:
\bbtitle{Estimations and Tests in Change-Point Models}.
\bpublisher{World Scientific Publishing Co. Pte Ltd.},
\blocation{London}
(\byear{2018}).
\doiurl{10.1142/10757}.
\burl{https://www.worldscientific.com/doi/abs/10.1142/10757}
\end{bbook}
\endbibitem

\bibitem{pageContinuousInspectionSchemes1954a}
\begin{barticle}
\bauthor{\bsnm{Page}, \binits{E.S.}}:
\batitle{Continuous {Inspection} {Schemes}}.
\bjtitle{Biometrika}
\bvolume{41}(\bissue{1/2}),
\bfpage{100}
(\byear{1954}).
\doiurl{10.2307/2333009}.
Accessed 2021-05-25
\end{barticle}
\endbibitem

\bibitem{wald}
\begin{barticle}
\bauthor{\bsnm{Wald}, \binits{A.}},
\bauthor{\bsnm{Wolfowitz}, \binits{J.}}:
\batitle{Optimum {Character} of the {Sequential} {Probability} {Ratio} {Test}}.
\bjtitle{The Annals of Mathematical Statistics}
\bvolume{19}(\bissue{3}),
\bfpage{326}--\blpage{339}
(\byear{1948})
\end{barticle}
\endbibitem

\bibitem{lord_1977}
\begin{botherref}
\oauthor{\bsnm{Lorden}, \binits{G.}}:
Procedures for reacting to a change in distribution.
The Annals of Statistics
\textbf{5}
(1977)
\end{botherref}
\endbibitem

\bibitem{mous_2004}
\begin{botherref}
\oauthor{\bsnm{Moustakides}, \binits{G.V.}}:
Optimality of the cusum procedure in continuous time.
The Annals of Statistics 2004-feb vol. 32 iss. 1
\textbf{32}
(2004).
\doiurl{10.2307/3448511}
\end{botherref}
\endbibitem

\bibitem{ritov}
\begin{botherref}
\oauthor{\bsnm{Ritov}, \binits{Y.}}:
Decision theoretic optimality of the cusum procedure.
The Annals of Statistics 1990-sep vol. 18 iss. 3
\textbf{18}
(1990).
\doiurl{10.1214/aos/1176347761}
\end{botherref}
\endbibitem

\bibitem{aue}
\begin{botherref}
\oauthor{\bsnm{Aue}, \binits{A.}},
\oauthor{\bsnm{Horváth}, \binits{L.}}:
Structural breaks in time series.
Journal of Time Series Analysis 2012-sep 14 vol. 34 iss. 1
\textbf{34}
(2012).
\doiurl{10.1111/j.1467-9892.2012.00819.x}
\end{botherref}
\endbibitem

\bibitem{Yang_Heart}
\begin{barticle}
\bauthor{\bsnm{Yang}, \binits{P.}},
\bauthor{\bsnm{Dumont}, \binits{G.}},
\bauthor{\bsnm{Ansermino}, \binits{J.M.}}:
\batitle{Adaptive change detection in heart rate trend monitoring in anesthetized children}.
\bjtitle{IEEE Transactions on Biomedical Engineering}
\bvolume{53},
\bfpage{2211}--\blpage{2219}
(\byear{2006}).
\doiurl{10.1109/tbme.2006.877107}
\end{barticle}
\endbibitem

\bibitem{harrison_Demand}
\begin{barticle}
\bauthor{\bsnm{Harrison}, \binits{P.J.}},
\bauthor{\bsnm{Davies}, \binits{O.L.}}:
\batitle{The use of cumulative sum (cusum) techniques for the control of routine forecasts of product demand}.
\bjtitle{Operations Research}
\bvolume{12},
\bfpage{325}--\blpage{333}
(\byear{1964}).
\doiurl{10.1287/opre.12.2.325}
\end{barticle}
\endbibitem

\bibitem{manner_Copula}
\begin{botherref}
\oauthor{\bsnm{Manner}, \binits{H.}},
\oauthor{\bsnm{Stark}, \binits{F.}},
\oauthor{\bsnm{Wied}, \binits{D.}}:
Testing for structural breaks in factor copula models.
Journal of Econometrics,
0304407618301842
(2018).
\doiurl{10.1016/j.jeconom.2018.10.001}
\end{botherref}
\endbibitem

\bibitem{burgEvaluationChangePoint2020}
\begin{botherref}
\oauthor{\bsnm{Burg}, \binits{G.J.J.v.d.}},
\oauthor{\bsnm{Williams}, \binits{C.K.I.}}:
An {Evaluation} of {Change} {Point} {Detection} {Algorithms}.
arXiv:2003.06222 [cs, stat]
(2020).
arXiv: 2003.06222.
Accessed 2022-01-14
\end{botherref}
\endbibitem

\bibitem{amin_2017}
\begin{botherref}
\oauthor{\bsnm{Aminikhanghahi}, \binits{D.J.} \bsuffix{Samaneh;~Cook}}:
A survey of methods for time series change point detection.
Knowledge and Information Systems 2016-sep 08 vol. 51 iss. 2
\textbf{51}
(2016).
\doiurl{10.1007/s10115-016-0987-z}
\end{botherref}
\endbibitem

\bibitem{siegmund}
\begin{bbook}
\bauthor{\bsnm{Siegmund}, \binits{D.}}:
\bbtitle{Sequential Analysis}.
\bpublisher{Springer},
\blocation{Heidelberg}
(\byear{1985})
\end{bbook}
\endbibitem

\bibitem{truong}
\begin{botherref}
\oauthor{\bsnm{Truong}, \binits{C.}},
\oauthor{\bsnm{Oudre}, \binits{L.}},
\oauthor{\bsnm{Vayatis}, \binits{N.}}:
Selective review of offline change point detection methods.
Signal Processing 2020-feb vol. 167
\textbf{167}
(2020).
\doiurl{10.1016/j.sigpro.2019.107299}
\end{botherref}
\endbibitem

\bibitem{cao}
\begin{botherref}
\oauthor{\bsnm{Cao}, \binits{Y.}},
\oauthor{\bsnm{Xie}, \binits{L.}},
\oauthor{\bsnm{Xie}, \binits{Y.}},
\oauthor{\bsnm{Xu}, \binits{H.}}:
Sequential change-point detection via online convex optimization.
Entropy 2018-feb 07 vol. 20 iss. 2
\textbf{20}
(2018).
\doiurl{10.3390/e20020108}
\end{botherref}
\endbibitem

\bibitem{hawkinsCumulativeSumCharts1998}
\begin{bbook}
\bauthor{\bsnm{Hawkins}, \binits{D.M.}},
\bauthor{\bsnm{Olwell}, \binits{D.H.}}:
\bbtitle{Cumulative {Sum} {Charts} and {Charting} for {Quality} {Improvement}}.
\bpublisher{Springer},
\blocation{New York, NY}
(\byear{1998}).
\doiurl{10.1007/978-1-4612-1686-5}.
\burl{http://link.springer.com/10.1007/978-1-4612-1686-5}
Accessed 2021-05-25
\end{bbook}
\endbibitem

\bibitem{lim}
\begin{botherref}
\oauthor{\bsnm{Abdul Halim~Lim}, \binits{S.}},
\oauthor{\bsnm{Antony}, \binits{J.}},
\oauthor{\bsnm{Arshed}, \binits{N.}},
\oauthor{\bsnm{Albliwi}, \binits{S.}}:
A systematic review of statistical process control implementation in the food manufacturing industry.
Total Quality Management \& Business Excellence,
1--14
(2015).
\doiurl{10.1080/14783363.2015.1050181}
\end{botherref}
\endbibitem

\bibitem{pan_2005}
\begin{barticle}
\bauthor{\bsnm{Tsiamyrtzis}, \binits{P.}},
\bauthor{\bsnm{Hawkins}, \binits{D.M.}}:
\batitle{A bayesian scheme to detect changes in the mean of a short-run process}.
\bjtitle{Technometrics}
\bvolume{47},
\bfpage{446}--\blpage{456}
(\byear{2005}).
\doiurl{10.2307/25471069}
\end{barticle}
\endbibitem

\bibitem{huwel_dynamically_2022}
\begin{bchapter}
\bauthor{\bsnm{H{\"u}wel}, \binits{J.D.}},
\bauthor{\bsnm{Haselbeck}, \binits{F.}},
\bauthor{\bsnm{Grimm}, \binits{D.G.}},
\bauthor{\bsnm{Beecks}, \binits{C.}}:
\bctitle{Dynamically {Self}-adjusting {Gaussian} {Processes} for {Data} {Stream} {Modelling}},
\bconflocation{Cham},
pp. \bfpage{96}--\blpage{114}
(\byear{2022})
\end{bchapter}
\endbibitem

\bibitem{ma_rul_2021}
\begin{barticle}
\bauthor{\bsnm{Ma}, \binits{Q.}},
\bauthor{\bsnm{Zheng}, \binits{Y.}},
\bauthor{\bsnm{Yang}, \binits{W.}},
\bauthor{\bsnm{Zhang}, \binits{Y.}},
\bauthor{\bsnm{Zhang}, \binits{H.}}:
\batitle{Remaining useful life prediction of lithium battery based on capacity regeneration point detection}.
\bjtitle{Energy}
\bvolume{234},
\bfpage{121233}
(\byear{2021}).
\doiurl{10.1016/j.energy.2021.121233}
\end{barticle}
\endbibitem

\bibitem{SQC}
\begin{bbook}
\bauthor{\bsnm{Montgomery}, \binits{D.C.}}:
\bbtitle{Statistical Quality Control},
\bedition{7}th edn.
\bpublisher{Wiley},
\blocation{Hoboken, New Jersey}
(\byear{2012})
\end{bbook}
\endbibitem

\bibitem{buecher}
\begin{botherref}
\oauthor{\bsnm{B\"{u}cher}, \binits{A.}},
\oauthor{\bsnm{Dette}, \binits{H.}},
\oauthor{\bsnm{Heinrichs}, \binits{F.}}:
Are deviations in a gradually varying mean relevant? A testing approach based on sup-norm estimators
(2020)
\end{botherref}
\endbibitem

\bibitem{vogt}
\begin{barticle}
\bauthor{\bsnm{Vogt}, \binits{M.}},
\bauthor{\bsnm{Dette}, \binits{H.}}:
\batitle{Detecting gradual changes in locally stationary processes}.
\bjtitle{The Annals of Statistics}
\bvolume{43},
\bfpage{713}--\blpage{740}
(\byear{2015}).
\doiurl{10.1214/14-aos1297}
\end{barticle}
\endbibitem

\bibitem{aueSteinbach}
\begin{barticle}
\bauthor{\bsnm{Aue}, \binits{A.}},
\bauthor{\bsnm{Steinebach}, \binits{J.}}:
\batitle{A note on estimating the change-point of a gradually changing stochastic process}.
\bjtitle{Statistics \& Probability Letters}
\bvolume{56},
\bfpage{177}--\blpage{191}
(\byear{2002}).
\doiurl{10.1016/s0167-7152(01)00184-5}
\end{barticle}
\endbibitem

\bibitem{woodallSTATISTICALDESIGNCUSUM1993}
\begin{barticle}
\bauthor{\bsnm{Woodall}, \binits{W.H.}},
\bauthor{\bsnm{Adams}, \binits{B.M.}}:
\batitle{{The} {statistical} {design} {of} {cusum} {charts}}.
\bjtitle{Quality Engineering}
\bvolume{5}(\bissue{4}),
\bfpage{559}--\blpage{570}
(\byear{1993}).
\doiurl{10.1080/08982119308918998}.
Accessed 2021-05-25
\end{barticle}
\endbibitem

\bibitem{bissell}
\begin{barticle}
\bauthor{\bsnm{Bissell}, \binits{A.F.}}:
\batitle{The performance of control charts and cusums under linear trend}.
\bjtitle{Journal of the Royal Statistical Society: Series C (Applied Statistics)}
\bvolume{33},
\bfpage{145}--\blpage{151}
(\byear{1984}).
\doiurl{10.2307/2347439}
\end{barticle}
\endbibitem

\bibitem{fearnhead_cpop_2019}
\begin{barticle}
\bauthor{\bsnm{Paul~Fearnhead}, \binits{R.M.}},
\bauthor{\bsnm{Letchford}, \binits{A.}}:
\batitle{Detecting changes in slope with an l0 penalty}.
\bjtitle{Journal of Computational and Graphical Statistics}
\bvolume{28}(\bissue{2}),
\bfpage{265}--\blpage{275}
(\byear{2019})
{\href{https://arxiv.org/abs/https://doi.org/10.1080/10618600.2018.1512868}{{https://doi.org/10.1080/10618600.2018.1512868}}}.
\doiurl{10.1080/10618600.2018.1512868}
\end{barticle}
\endbibitem

\bibitem{verbesselt_detecting_2010}
\begin{barticle}
\bauthor{\bsnm{Verbesselt}, \binits{J.}},
\bauthor{\bsnm{Hyndman}, \binits{R.}},
\bauthor{\bsnm{Newnham}, \binits{G.}},
\bauthor{\bsnm{Culvenor}, \binits{D.}}:
\batitle{Detecting trend and seasonal changes in satellite image time series}.
\bjtitle{Remote Sensing of Environment}
\bvolume{114}(\bissue{1}),
\bfpage{106}--\blpage{115}
(\byear{2010}).
\doiurl{10.1016/j.rse.2009.08.014}
\end{barticle}
\endbibitem

\bibitem{AdamsMacKay}
\begin{botherref}
\oauthor{\bsnm{Adams}, \binits{R.P.}},
\oauthor{\bsnm{MacKay}, \binits{D.J.C.}}:
Bayesian online changepoint detection
(2007)
\end{botherref}
\endbibitem

\bibitem{chen_OCD_2022}
\begin{barticle}
\bauthor{\bsnm{Chen}, \binits{Y.}},
\bauthor{\bsnm{Wang}, \binits{T.}},
\bauthor{\bsnm{Samworth}, \binits{R.J.}}:
\batitle{{High-Dimensional, Multiscale Online Changepoint Detection}}.
\bjtitle{Journal of the Royal Statistical Society Series B: Statistical Methodology}
\bvolume{84}(\bissue{1}),
\bfpage{234}--\blpage{266}
(\byear{2022}).
\doiurl{10.1111/rssb.12447}
\end{barticle}
\endbibitem

\bibitem{agudelo}
\begin{botherref}
\oauthor{\bsnm{Agudelo-Espa{\~n}a}, \binits{D.}},
\oauthor{\bsnm{Gomez-Gonzalez}, \binits{S.}},
\oauthor{\bsnm{Bauer}, \binits{S.}},
\oauthor{\bsnm{Sch\"{o}lkopf}, \binits{B.}},
\oauthor{\bsnm{Peters}, \binits{J.}}:
Bayesian {Online} {Prediction} of {Change} {Points}.
arXiv:1902.04524 [cs, stat]
(2020).
arXiv: 1902.04524.
Accessed 2022-05-09
\end{botherref}
\endbibitem

\bibitem{wang}
\begin{bchapter}
\bauthor{\bsnm{Wang}, \binits{Z.}},
\bauthor{\bsnm{Lin}, \binits{X.}},
\bauthor{\bsnm{Mishra}, \binits{A.}},
\bauthor{\bsnm{Sriharsha}, \binits{R.}}:
\bctitle{Online changepoint detection on a budget}.
In: \bbtitle{2021 International Conference on Data Mining Workshops (ICDMW)},
pp. \bfpage{414}--\blpage{420}.
\bpublisher{IEEE Computer Society},
\blocation{Los Alamitos, CA, USA}
(\byear{2021}).
\doiurl{10.1109/ICDMW53433.2021.00057}.
\burl{https://doi.ieeecomputersociety.org/10.1109/ICDMW53433.2021.00057}
\end{bchapter}
\endbibitem

\bibitem{gardner85}
\begin{barticle}
\bauthor{\bsnm{Gardner}, \binits{E.S.}},
\bauthor{\bsnm{Mckenzie}, \binits{E.}}:
\batitle{Forecasting {Trends} in {Time} {Series}}.
\bjtitle{Management Science}
\bvolume{31}(\bissue{10}),
\bfpage{1237}--\blpage{1246}
(\byear{1985}).
\doiurl{10.1287/mnsc.31.10.1237}.
Accessed 2023-08-17
\end{barticle}
\endbibitem

\bibitem{krause}
\begin{botherref}
\oauthor{\bsnm{Krause}, \binits{M.}}:
Unsupervised change point detection for heterogeneous sensor signals
(2023).
arXiv:2305.11976v1
\end{botherref}
\endbibitem

\bibitem{bukovsky2019}
\begin{botherref}
\oauthor{\bsnm{Bukovsky}, \binits{I.}},
\oauthor{\bsnm{Kinsner}, \binits{W.}},
\oauthor{\bsnm{Homma}, \binits{N.}}:
Learning entropy as a learning-based information concept.
Entropy
\textbf{21}(2)
(2019).
\doiurl{10.3390/e21020166}
\end{botherref}
\endbibitem

\bibitem{information15}
\begin{barticle}
\bauthor{\bsnm{Wiggins}, \binits{P.A.}}:
\batitle{An {Information}-{Based} {Approach} to {Change}-{Point} {Analysis} with {Applications} to {Biophysics} and {Cell} {Biology}}.
\bjtitle{Biophysical Journal}
\bvolume{109}(\bissue{2}),
\bfpage{346}--\blpage{354}
(\byear{2015}).
\doiurl{10.1016/j.bpj.2015.05.038}
\end{barticle}
\endbibitem

\bibitem{foo}
\begin{botherref}
\oauthor{\bsnm{Yu}, \binits{H.}},
\oauthor{\bsnm{Liu}, \binits{T.}},
\oauthor{\bsnm{Lu}, \binits{J.}},
\oauthor{\bsnm{Zhang}, \binits{G.}}:
Automatic Learning to Detect Concept Drift.
arXiv
(2021).
\doiurl{10.48550/ARXIV.2105.01419}.
\url{https://arxiv.org/abs/2105.01419}
\end{botherref}
\endbibitem

\bibitem{SemiSupervised}
\begin{bchapter}
\bauthor{\bsnm{De~Brabandere}, \binits{A.}},
\bauthor{\bsnm{Cao}, \binits{Z.}},
\bauthor{\bsnm{De~Vos}, \binits{M.}},
\bauthor{\bsnm{Bertrand}, \binits{A.}},
\bauthor{\bsnm{Davis}, \binits{J.}}:
\bctitle{Semi-supervised change point detection using active learning}.
In: \bbtitle{Discovery Science: 25th International Conference, DS 2022, Montpellier, France, October 10–12, 2022, Proceedings},
pp. \bfpage{74}--\blpage{88}.
\bpublisher{Springer},
\blocation{Berlin, Heidelberg}
(\byear{2022}).
\doiurl{10.1007/978-3-031-18840-4_6}.
\burl{https://doi.org/10.1007/978-3-031-18840-4_6}
\end{bchapter}
\endbibitem

\bibitem{Fearnhead2007}
\begin{barticle}
\bauthor{\bsnm{Liu}, \binits{P.F.Z.}}:
\batitle{On-line inference for multiple changepoint problems}.
\bjtitle{Journal Of The Royal Statistical Society}
\bvolume{69},
\bfpage{589}--\blpage{605}
(\byear{2007}).
\doiurl{10.1111/j.1467-9868.2007.00601.x}
\end{barticle}
\endbibitem

\bibitem{diego_2020}
\begin{botherref}
\oauthor{\bsnm{Agudelo-Espa{\~n}a}, \binits{D.}},
\oauthor{\bsnm{Gomez-Gonzalez}, \binits{S.}},
\oauthor{\bsnm{Bauer}, \binits{S.}},
\oauthor{\bsnm{Sch\"{o}lkopf}, \binits{B.}},
\oauthor{\bsnm{Peters}, \binits{J.}}:
Bayesian {Online} {Prediction} of {Change} {Points}.
arXiv:1902.04524 [cs, stat]
(2020).
arXiv: 1902.04524.
Accessed 2022-05-09
\end{botherref}
\endbibitem

\bibitem{malladi}
\begin{bchapter}
\bauthor{\bsnm{Malladi}, \binits{R.}},
\bauthor{\bsnm{Kalamangalam}, \binits{G.P.}},
\bauthor{\bsnm{Aazhang}, \binits{B.}}:
\bctitle{Online {Bayesian} change point detection algorithms for segmentation of epileptic activity}.
In: \bbtitle{2013 {Asilomar} {Conference} on {Signals}, {Systems} and {Computers}},
pp. \bfpage{1833}--\blpage{1837}.
\bpublisher{IEEE},
\blocation{Pacific Grove, CA, USA}
(\byear{2013}).
\doiurl{10.1109/ACSSC.2013.6810619}.
\burl{http://ieeexplore.ieee.org/document/6810619/}
Accessed 2023-08-29
\end{bchapter}
\endbibitem

\bibitem{pan_2022}
\begin{barticle}
\bauthor{\bsnm{Konstantinos~Bourazas}, \binits{D.K.}},
\bauthor{\bsnm{Tsiamyrtzis}, \binits{P.}}:
\batitle{Predictive {Control} {Charts} ({PCC}): {A} {Bayesian} approach in online monitoring of short runs}.
\bjtitle{Journal of Quality Technology}
\bvolume{54}(\bissue{4}),
\bfpage{367}--\blpage{391}
(\byear{2022}).
\doiurl{10.1080/00224065.2021.1916413}.
\bcomment{Publisher: Taylor \& Francis \_eprint: https://doi.org/10.1080/00224065.2021.1916413}
\end{barticle}
\endbibitem

\bibitem{lau}
\begin{bchapter}
\bauthor{\bsnm{Lau}, \binits{H.F.}},
\bauthor{\bsnm{Yamamoto}, \binits{S.}}:
\bctitle{Bayesian online changepoint detection to improve transparency in human-machine interaction systems}.
In: \bbtitle{49th {IEEE} {Conference} on {Decision} and {Control} ({CDC})},
pp. \bfpage{3572}--\blpage{3577}.
\bpublisher{IEEE},
\blocation{Atlanta, GA, USA}
(\byear{2010}).
\doiurl{10.1109/CDC.2010.5717959}.
\burl{http://ieeexplore.ieee.org/document/5717959/}
Accessed 2023-08-29
\end{bchapter}
\endbibitem

\bibitem{survival}
\begin{bbook}
\bauthor{\bparticle{van} \bsnm{Houwelingen; Joseph G. Ibrahim; Thomas H.~Scheike}, \binits{J.P.K.H.C.}}:
\bbtitle{Handbook of Survival Analysis}.
\bsertitle{Chapman \& Hall / CRC Handbooks of Modern Statistical Methods}.
\bpublisher{Chapman and Hall/CRC},
\blocation{Boca Raton, FL}
(\byear{2013}).
\bcomment{{ISBN=}978-1-4665-5567-9,978-1-4665-5566-2}
\end{bbook}
\endbibitem

\bibitem{pan_2023}
\begin{barticle}
\bauthor{\bsnm{Konstantinos~Bourazas}, \binits{F.S.}},
\bauthor{\bsnm{Tsiamyrtzis}, \binits{P.}}:
\batitle{Design and properties of the predictive ratio cusum ({PRC}) control charts}.
\bjtitle{Journal of Quality Technology}
\bvolume{55}(\bissue{4}),
\bfpage{404}--\blpage{421}
(\byear{2023}).
\doiurl{10.1080/00224065.2022.2161435}.
\bcomment{Publisher: Taylor \& Francis \_eprint: https://doi.org/10.1080/00224065.2022.2161435}
\end{barticle}
\endbibitem

\bibitem{verbesselt_near_2012}
\begin{barticle}
\bauthor{\bsnm{Verbesselt}, \binits{J.}},
\bauthor{\bsnm{Zeileis}, \binits{A.}},
\bauthor{\bsnm{Herold}, \binits{M.}}:
\batitle{Near real-time disturbance detection using satellite image time series}.
\bjtitle{Remote Sensing of Environment}
\bvolume{123},
\bfpage{98}--\blpage{108}
(\byear{2012}).
\doiurl{10.1016/j.rse.2012.02.022}.
Accessed 2022-03-28
\end{barticle}
\endbibitem

\bibitem{shar_2016}
\begin{barticle}
\bauthor{\bsnm{Sharma}, \binits{S.}},
\bauthor{\bsnm{Swayne}, \binits{D.A.}},
\bauthor{\bsnm{Obimbo}, \binits{C.}}:
\batitle{Trend analysis and change point techniques: a survey}.
\bjtitle{Energy, Ecology and Environment}
\bvolume{1}(\bissue{3}),
\bfpage{123}--\blpage{130}
(\byear{2016}).
\doiurl{10.1007/s40974-016-0011-1}.
Accessed 2022-05-08
\end{barticle}
\endbibitem

\bibitem{bassevilleFAULTISOLATIONDIAGNOSIS2002}
\begin{botherref}
\oauthor{\bsnm{Basseville}, \binits{M.}},
\oauthor{\bsnm{Nikiforov}, \binits{I.}}:
{Fault} {isolation} {for} {diagnosis}: {nuisance} {rejection} {and} {multiple} {hypotheses} {testing}.
Annual Reviews in Control,
14
(2002)
\end{botherref}
\endbibitem

\bibitem{bissell2}
\begin{barticle}
\bauthor{\bsnm{Bissell}, \binits{A.F.}}:
\batitle{Cusum techniques for quality control (with discussion)}.
\bjtitle{Applied Statistics}
\bvolume{18},
\bfpage{1}--\blpage{30}
(\byear{1969})
\end{barticle}
\endbibitem

\bibitem{variance}
\begin{barticle}
\bauthor{\bsnm{Chang}, \binits{T.C.}},
\bauthor{\bsnm{Gan}, \binits{F.F.}}:
\batitle{A cumulative sum control chart for monitoring process variance}.
\bjtitle{Journal of Quality Technology}
\bvolume{27}(\bissue{2}),
\bfpage{109}--\blpage{119}
(\byear{1995})
{\href{https://arxiv.org/abs/https://doi.org/10.1080/00224065.1995.11979574}{{https://doi.org/10.1080/00224065.1995.11979574}}}.
\doiurl{10.1080/00224065.1995.11979574}
\end{barticle}
\endbibitem

\bibitem{jech_radionuclide_2018}
\begin{bchapter}
\bauthor{\bsnm{Jech}, \binits{M.}},
\bauthor{\bsnm{Lenauer}, \binits{C.}}:
\bctitle{Radionuclide methods}.
In: \bbtitle{Friction, Lubrication, and Wear Technology}.
\bsertitle{{ASM} Handbook},
vol. \bseriesno{18},
pp. \bfpage{1045}--\blpage{1055}.
\bpublisher{{ASM} International},
\blocation{Ohio, USA}
(\byear{2017})
\end{bchapter}
\endbibitem

\bibitem{AC1}
\begin{botherref}
\oauthor{\bsnm{Glock}, \binits{A.-C.}},
\oauthor{\bsnm{Sobieczky}, \binits{F.}},
\oauthor{\bsnm{Jech}, \binits{M.}}:
Detection of anomalous events in the wear-behaviour of continuously recorded sliding friction pairs.
Conference Proceedings \"{O}TG-Tagung 2019,
30--40
(2019).
{ISBN:}9783901657627
\end{botherref}
\endbibitem

\bibitem{Pstandard}
\begin{bbook}
\bauthor{\bsnm{Feller}, \binits{W.}}:
\bbtitle{Probability Theory}
vol. \bseriesno{II}.
\bpublisher{Wiley},
\blocation{Hoboken, New Jersey}
(\byear{2019}).
\bcomment{Chap. VI.6}
\end{bbook}
\endbibitem

\bibitem{bfastMosum_zeileis_2005}
\begin{barticle}
\bauthor{\bsnm{Zeileis}, \binits{A.}},
\bauthor{\bsnm{Leisch}, \binits{F.}},
\bauthor{\bsnm{Kleiber}, \binits{C.}},
\bauthor{\bsnm{Hornik}, \binits{K.}}:
\batitle{Monitoring structural change in dynamic econometric models}.
\bjtitle{Journal of Applied Econometrics}
\bvolume{20}(\bissue{1}),
\bfpage{99}--\blpage{121}
(\byear{2005}).
\doiurl{10.1002/jae.776}.
Accessed 2022-10-12
\end{barticle}
\endbibitem

\bibitem{brisset_radiotracer_2020}
\begin{bbook}
\bauthor{\bsnm{Brisset}, \binits{P.}},
\bauthor{\bsnm{Ditroi}, \binits{F.}},
\bauthor{\bsnm{Eberle}, \binits{D.}},
\bauthor{\bsnm{Jech}, \binits{M.}},
\bauthor{\bsnm{Kleinrahm}, \binits{A.}},
\bauthor{\bsnm{Lenauer}, \binits{C.}},
\bauthor{\bsnm{Sauvage}, \binits{T.}},
\bauthor{\bsnm{Thereska}, \binits{J.}}:
\bbtitle{Radiotracer Technologies for Wear, Erosion and Corrosion Measurement}.
\bsertitle{TECDOC Series},
vol. \bseriesno{1897}.
\bpublisher{International Atomic Energy Agency},
\blocation{Vienna}
(\byear{2020}).
\bcomment{Chap. 5.5.3}
\end{bbook}
\endbibitem

\bibitem{qcc_2004}
\begin{barticle}
\bauthor{\bsnm{Scrucca}, \binits{L.}}:
\batitle{qcc: an r package for quality control charting and statistical process control}.
\bjtitle{R News}
\bvolume{4/1},
\bfpage{11}--\blpage{17}
(\byear{2004})
\end{barticle}
\endbibitem

\bibitem{forecast_2008}
\begin{barticle}
\bauthor{\bsnm{Hyndman}, \binits{R.J.}},
\bauthor{\bsnm{Khandakar}, \binits{Y.}}:
\batitle{Automatic time series forecasting: the forecast package for {R}}.
\bjtitle{Journal of Statistical Software}
\bvolume{26}(\bissue{3}),
\bfpage{1}--\blpage{22}
(\byear{2008}).
\doiurl{10.18637/jss.v027.i03}
\end{barticle}
\endbibitem

\bibitem{forecast_2022}
\begin{botherref}
\oauthor{\bsnm{Hyndman}, \binits{R.}},
\oauthor{\bsnm{Athanasopoulos}, \binits{G.}},
\oauthor{\bsnm{Bergmeir}, \binits{C.}},
\oauthor{\bsnm{Caceres}, \binits{G.}},
\oauthor{\bsnm{Chhay}, \binits{L.}},
\oauthor{\bsnm{O'Hara-Wild}, \binits{M.}},
\oauthor{\bsnm{Petropoulos}, \binits{F.}},
\oauthor{\bsnm{Razbash}, \binits{S.}},
\oauthor{\bsnm{Wang}, \binits{E.}},
\oauthor{\bsnm{Yasmeen}, \binits{F.}}:
{forecast}: Forecasting Functions for Time Series and Linear Models.
(2022).
R package version 8.16.
\url{https://pkg.robjhyndman.com/forecast/}
\end{botherref}
\endbibitem

\bibitem{kerasR_2017}
\begin{botherref}
\oauthor{\bsnm{Chollet}, \binits{F.}},
\oauthor{\bsnm{Allaire}, \binits{J.}}, et al.:
R Interface to Keras.
GitHub
(2017)
\end{botherref}
\endbibitem

\bibitem{hinkley70}
\begin{barticle}
\bauthor{\bsnm{Hinkley}, \binits{D.V.}}:
\batitle{Inference about the change-point in a sequence of random variables}.
\bjtitle{Biometrika}
\bvolume{57},
\bfpage{1}--\blpage{17}
(\byear{1970}).
\doiurl{10.2307/2334932}
\end{barticle}
\endbibitem

\end{thebibliography}
\end{document}